\documentclass[10pt,twocolumn,letterpaper]{article}

\usepackage{cvpr}
\usepackage{times}
\usepackage{epsfig}
\usepackage{graphicx}
\usepackage{amsmath}
\usepackage{amssymb}

\usepackage{caption}
\usepackage{subcaption}
\usepackage{algorithm}
\usepackage{algpseudocode}

\usepackage[overlay]{textpos}

\usepackage[pagebackref=true,breaklinks=true,letterpaper=true,colorlinks,bookmarks=false]{hyperref}

\cvprfinalcopy 

\def\cvprPaperID{2137} 

\ifcvprfinal\fi
\begin{document}

\title{A Genetic Algorithm-Based Solver for Very Large Jigsaw Puzzles}

\author{Dror Sholomon\\
{\tt\small dror.sholomon@gmail.com}
\and
Eli (Omid) David\thanks{www.elidavid.com}\\
{\tt\small mail@elidavid.com}
\and
Nathan S. Netanyahu\thanks{Nathan Netanyahu is also affiliated with the Center for Automation Research, University of Maryland, College Park, MD 20742 (e-mail: nathan@cfar.umd.edu).}\\
{\tt\small nathan@cs.biu.ac.il}
\and
\\
Department of Computer Science, Bar-Ilan University, Ramat-Gan 52900, Israel\\
}

\maketitle

\begin{textblock*}{10in}(-3mm, -65mm)
{\textbf{Ref:} \emph{IEEE Conference on Computer Vision and Pattern Recognition (CVPR)}, pages 1767--1774, Portland, OR, June 2013.}
\end{textblock*}

\begin{abstract}
   In this paper we propose the first effective automated, genetic algorithm (GA)-based jigsaw puzzle solver. We introduce a novel procedure of merging two "parent" solutions to an improved "child" solution by detecting, extracting, and combining correctly assembled puzzle segments. The solver proposed exhibits state-of-the-art performance solving previously attempted puzzles faster and far more accurately, and also puzzles of size never before attempted. Other contributions include the creation of a benchmark of large images, previously unavailable. We share the data sets and all of our results for future testing and comparative evaluation of jigsaw puzzle solvers.
\end{abstract}

\section{Introduction}
The problem domain of jigsaw puzzles is widely known to almost every human being from childhood. Given $n$ different non-overlapping pieces of an image, the player has to reconstruct the original image, taking advantage of both the shape and chromatic information of each piece. Although this popular game was proven to be an NP-complete problem ~\cite{journals/aai/Altman89}~\cite{springerlink:10.1007/s00373-007-0713-4}, it has been played successfully by children worldwide. Solutions to this problem might benefit the fields of biology~\cite{journals/science/MarandeB07}, chemistry~\cite{oai:xtcat.oclc.org:OCLCNo/ocm45147791}, literature~\cite{conf/ifip/MortonL68}, speech descrambling~\cite{Zhao:2007:PSA:1348258.1348289}, archeology~\cite{journals/tog/BrownTNBDVDRW08}~\cite{journals/KollerL06}, image editing~\cite{bb43059} and the recovery of shredded documents or photographs~\cite{cao2010automated}~\cite{marques2009reconstructing}~\cite{justino2006reconstructing}~\cite{conf/icip/DeeverG12}. Besides, as Goldberg {\etal}~\cite{GolMalBer04} have noted, the jigsaw puzzle problem may and should be researched for the sole reason that it stirs pure interest.

Jigsaw puzzles were first produced around 1760 by John Spilsbury, a Londonian engraver and mapmaker. Nevertheless, the first attempt by the scientific community to computationally solve the problem is attributed to Freeman and Garder~\cite{bb47278} who in 1964 introduced a solver which could handle up to nine-piece problems. Ever since then, the research focus regarding the problem has shifted from shape-based to merely color-based solvers of square-tile puzzles. In 2010 Cho {\etal}~\cite{conf/cvpr/ChoAF10} presented a probabilistic puzzle solver that could handle up to 432 pieces, given some a priori knowledge of the puzzle. Their results were improved a year later by Yang {\etal}~\cite{yang2011particle} who presented a particle filter-based solver. Furthermore, Pomeranz {\etal}~\cite{conf/cvpr/PomeranzSB11} introduced that year, for the first time, a fully automated square jigsaw puzzle solver that could handle puzzles of up to 3,000 pieces. Gallagher~\cite{conf/cvpr/Gallagher12} has advanced this even further by considering a more general variant of the problem, where neither piece orientation nor puzzle dimensions are known.

\begin{figure}
\centering
         \begin{subfigure}[t]{0.20\textwidth}
                \centering
                \includegraphics[width=\textwidth]{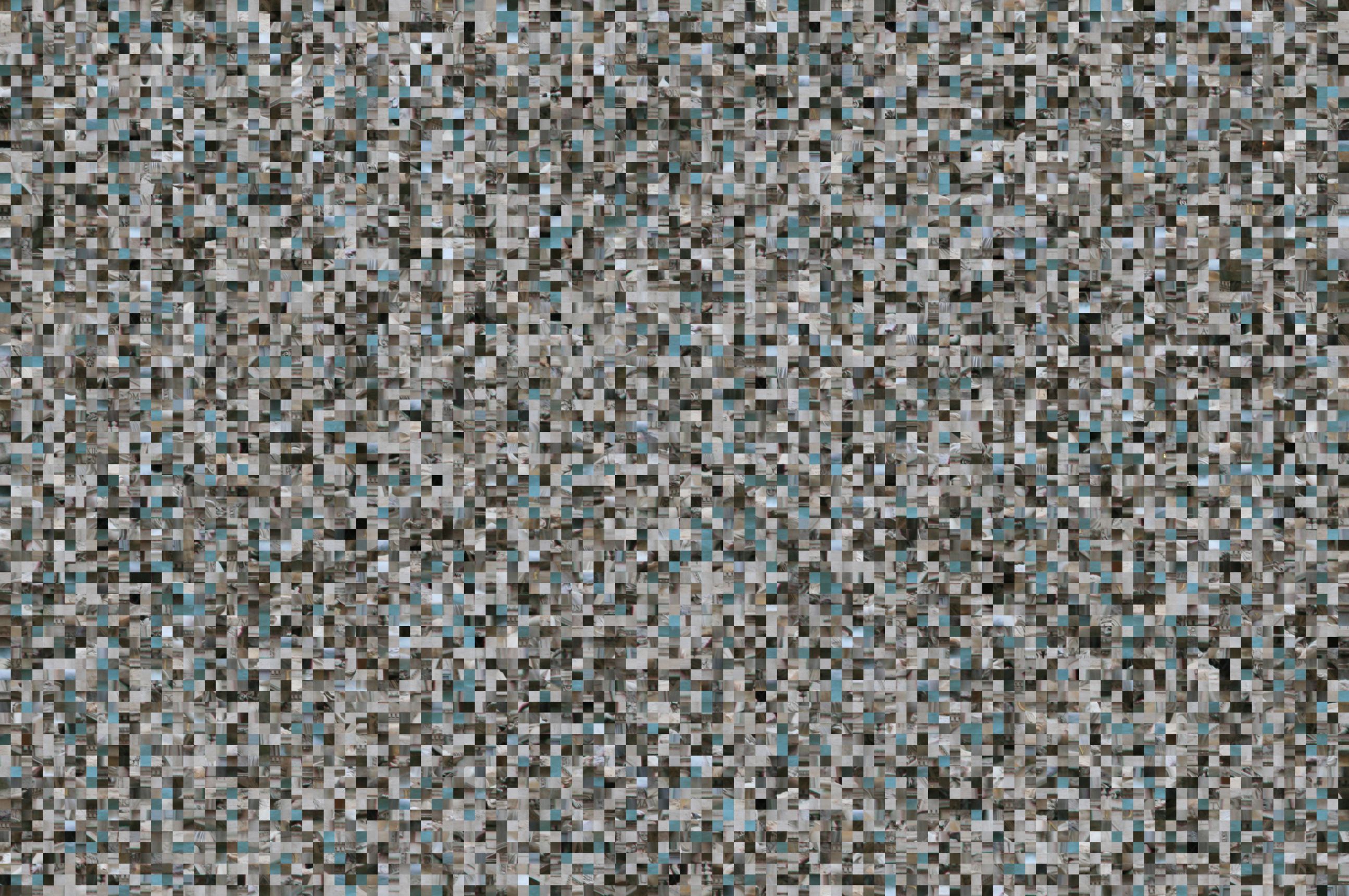}
                \caption{}
                \label{fig:intro_10375_gen_00000000}
        \end{subfigure}
        ~
        \begin{subfigure}[t]{0.20\textwidth}
                \centering
                \includegraphics[width=\textwidth]{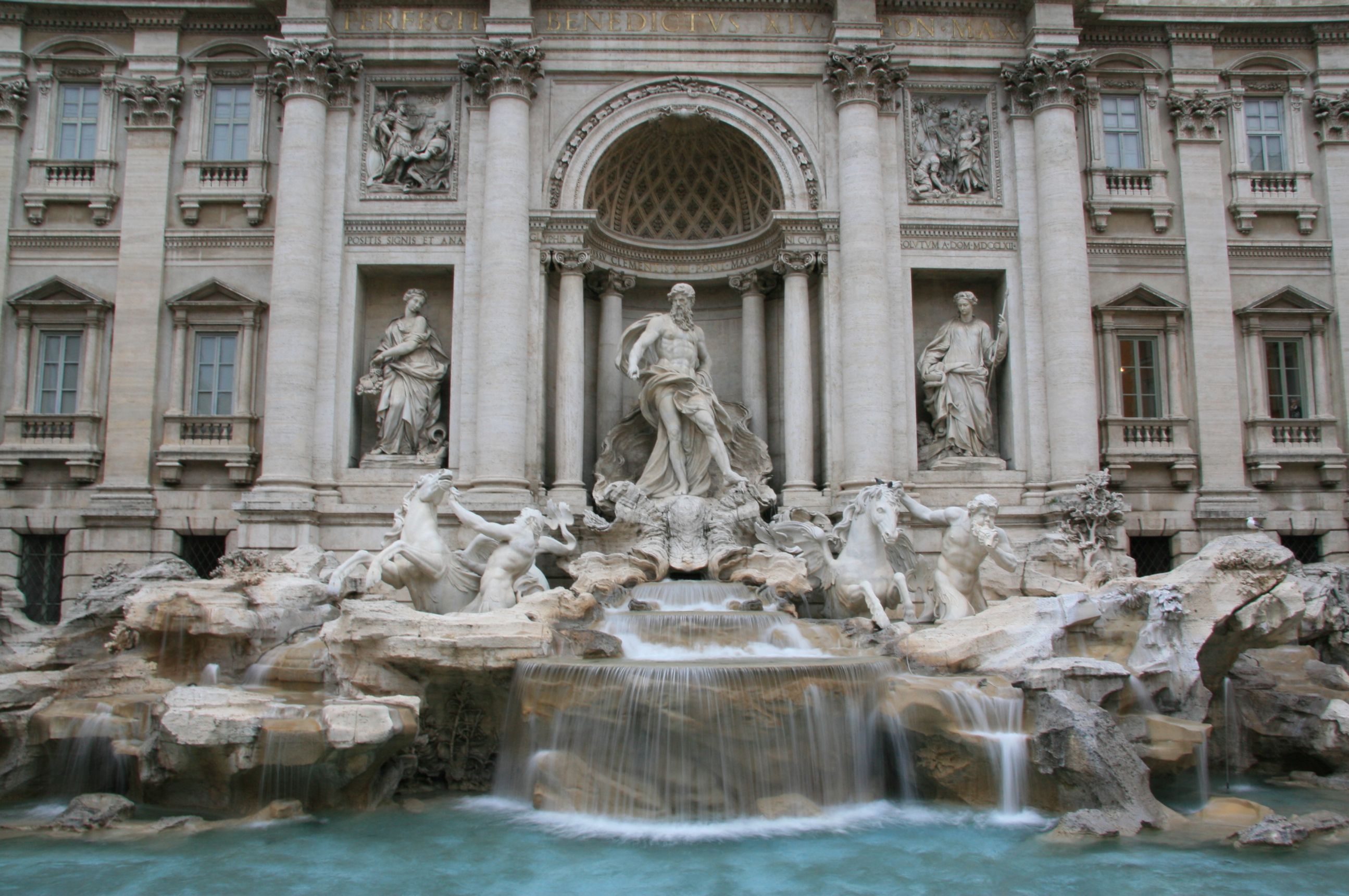}
                \caption{}
                \label{fig:intro_10375_orig}
        \end{subfigure}
	\begin{subfigure}[t]{0.20\textwidth}
                \centering
                \includegraphics[width=\textwidth]{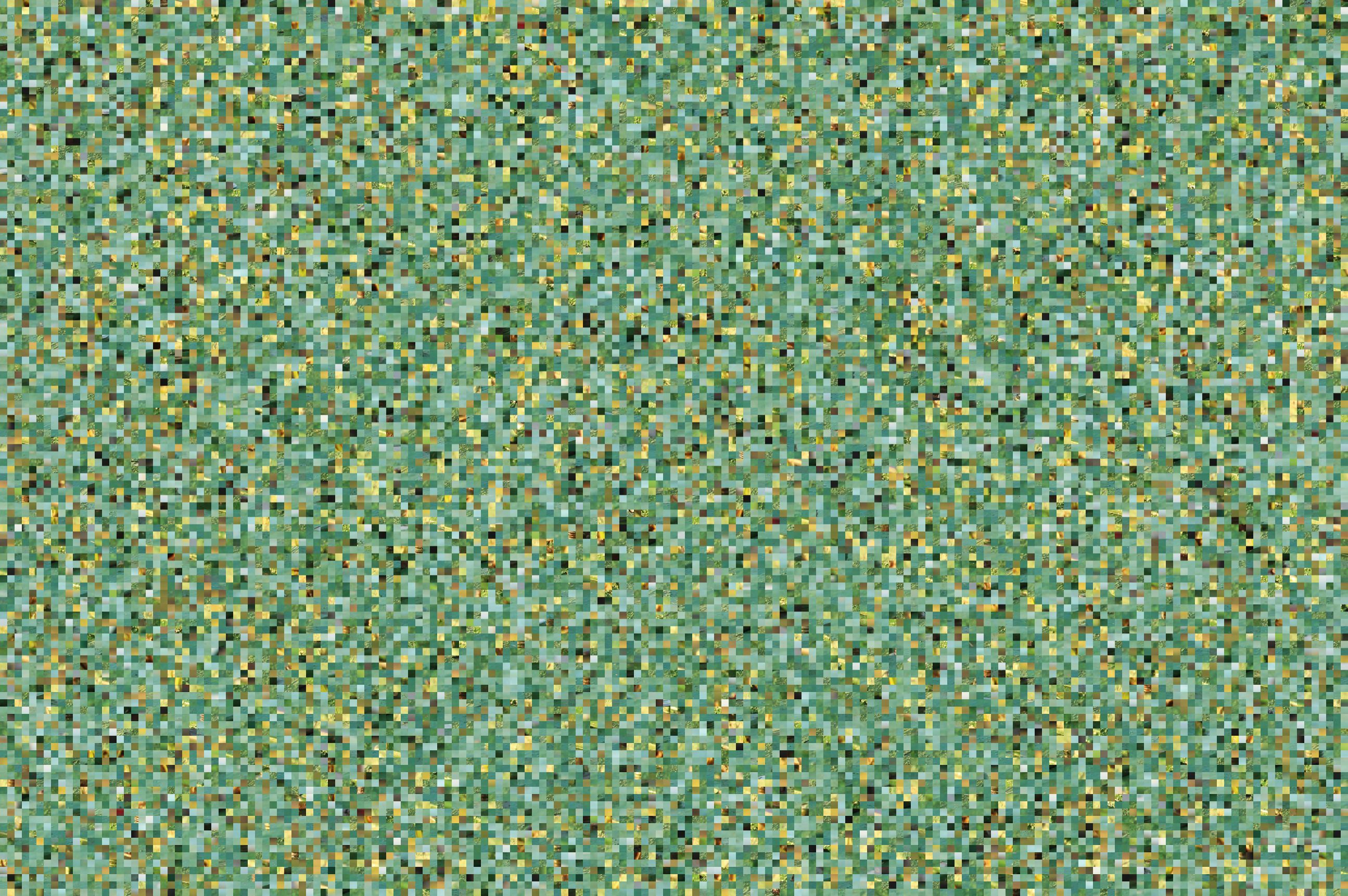}
                \caption{}
                \label{fig:intro_22834_gen_00000000}
        \end{subfigure}%
        ~ 
        \begin{subfigure}[t]{0.20\textwidth}
                \centering
                \includegraphics[width=\textwidth]{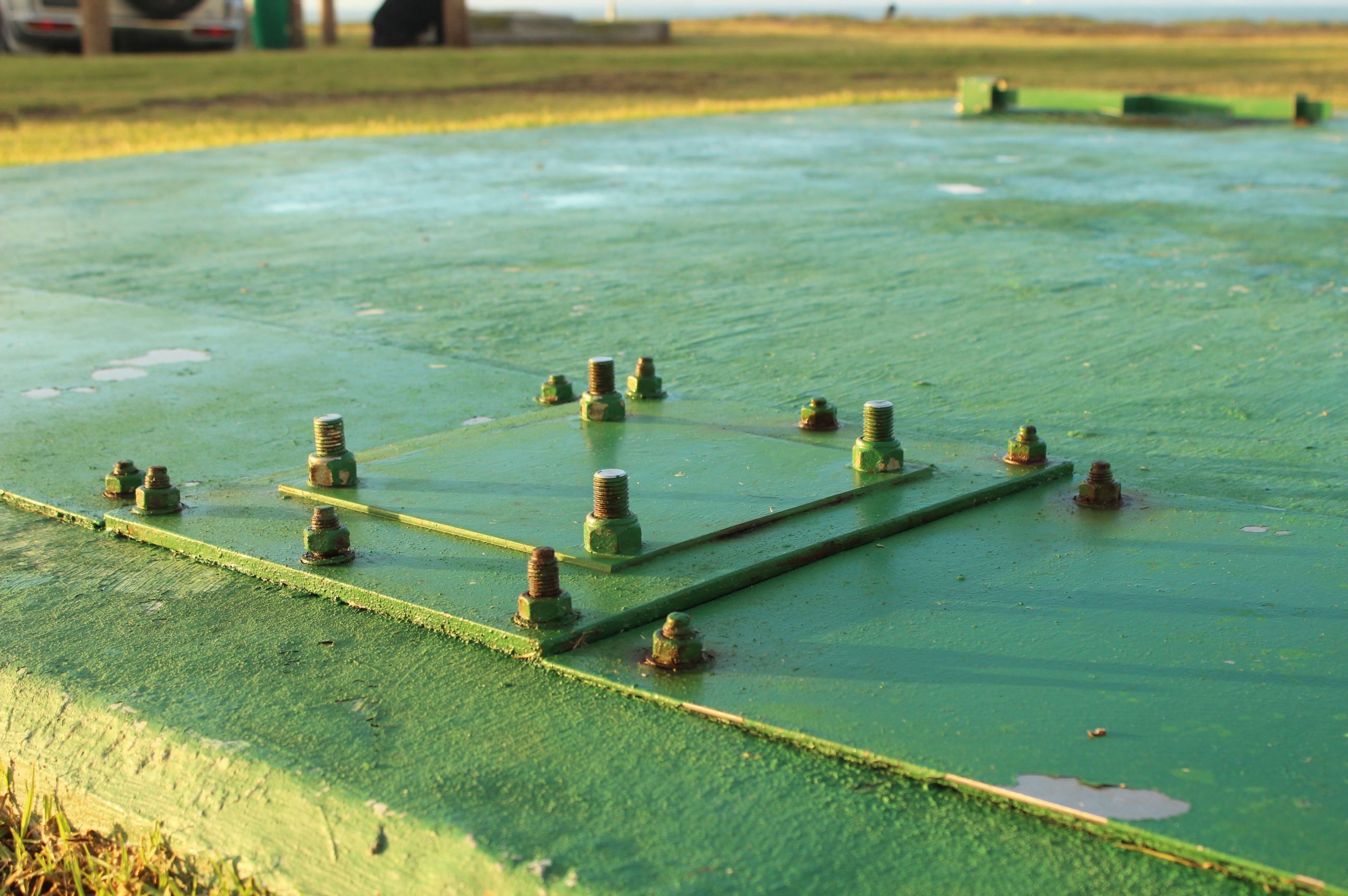}
                \caption{}
                \label{fig:intro_22834_gen_00000100}
        \end{subfigure}
        \caption{ Jigsaw puzzles before and after reassembly using our genetic algorithm-based solver. We believe these puzzles, of 10,375 (a-b) and 22,834 pieces (c-d), to be the largest automatically solved to date. }       
        \label{fig:introFig}
\end{figure}

In its most basic form, every puzzle solver requires an estimation function to evaluate the compatibility of adjacent pieces and a strategy for placing the pieces (as accurately as possible). Although much effort has been invested in perfecting the compatibility functions, recent strategies tend to be greedy, which is known to be problematic when encountering local optima. Thus, despite achieving very good (if not perfect) solutions for some puzzles, supplementary materials provided by Pomeranz {\etal}~\cite{conf/cvpr/site/PomeranzSB11} indicate that there is much room for improvement for many other puzzles. Comparative studies conducted by Gallagher (\cite{conf/cvpr/Gallagher12}, Table 4), regarding the benchmark set of 432-piece images, reveal only a slight improvement in accuracy relatively to Pomeranz {\etal} (95.1\% vs. 95.0\%). To the best of our knowledge, no additional benchmark runs have been reported by Gallagher. We thus assume that his method's performance on other benchmarks is comparable to that reported by Pomeranz {\etal} Interestingly, despite the availability of puzzle solvers for 3,000- and 9,000-piece puzzles, there exists no image set, for the purpose of benchmark testing, containing puzzles with more then 805 pieces. Current state-of-the-art solvers were only run on very few large images. Furthermore, these images were admittedly considered "easier" for solving~\cite{conf/cvpr/Gallagher12}, containing an extreme variety of textures and colors. We assume that similarly to the case of the smaller images, the accuracy of current solvers on some large puzzles could be greatly increased by using more sophisticated algorithms.

In this paper we harness the powerful technique of {\em genetic algorithms (GAs)}~\cite{holland1975adaptation} as a strategy for piece placement. The design of a GA-based solver has been attempted by Toyama {\etal}~\cite{bb58987}, but its successful performance was limited to 64-piece puzzles. We offer three major contributions. First and foremost, we present a significantly more accurate solver of the original jigsaw variant with known piece orientation and puzzle dimensions. Our solver compromises neither speed nor size as it outperforms state-of-the-art solvers, successfully tackling up to 22,834-piece size puzzles (more than twice the number of pieces ever attempted/reported) within a reasonable time frame. (See Figure~\ref{fig:introFig}.) Secondly, we assemble a new benchmark, consisting of sets of larger images (with varying degrees of difficulty), which we make public to the community~\cite{conf/cvpr/site/Our}. Also, we share all of our results (on this benchmark and other public datasets) for future testing and comparative evaluation of jigsaw puzzle solvers. Finally, we provide for the first time an effective GA-based puzzle solver, which should benefit research regarding the area of evolutionary computation (EC), in general, and the jigsaw puzzle problem, in particular. From an EC perspective, our novel techniques could be used for solving additional problems with similar properties. As to the jigsaw puzzle problem, our proposed framework could prove useful for solving more advanced variants, such as puzzles with missing pieces, unknown piece orientation, and more.

\section{Genetic algorithms}
A GA is a search procedure inside a problem's solution domain. Since examining all possible solutions of a specific problem is usually considered infeasible, GAs offer an optimization heuristic inspired by the theory of natural selection.

First, an initial {\em population} of candidate solutions, also called {\em chromosomes}, is randomly generated. Every chromosome is a complete solution to the problem, {\eg} a suggested arrangement of the puzzle's pieces. Next, various biologically inspired operators such as {\em selection}, {\em reproduction} and {\em mutation} are applied. These operators gradually improve the solutions in the population, eventually reaching the optimum solution ({\ie} the correct image).

In order to imitate natural selection, a chromosome's reproduction rate, {\ie} the number of times it is selected to reproduce and hence the number of its offsprings, is set directly proportionate to its {\em fitness}. The fitness is a score obtained by a {\em fitness function} and it represents the quality of a given solution. Thus, "good" solutions will have relatively more offsprings than other solutions. Moreover, good chromosomes are more likely to reproduce with other good chromosomes. The reproduction operator, called {\em crossover}, should allow the better traits from both parents to be passed on and be combined into the child solution, potentially creating an improved solution.

The success of a GA is mainly dependent on choosing an appropriate chromosome representation, crossover operator, and fitness function. The chromosome representation and crossover operator must allow the merge of two good solutions to an even better solution. The fitness function must correctly detect chromosomes containing promising solution parts to be passed on to the next generations.

\section{GA-based puzzle solver}
A basic GA framework for solving the jigsaw puzzle problem is given by the pseudocode of Algorithm~\ref{alg:GAMainLoop}. As previously noted, the GA contains a population of chromosomes, each of which represents a possible solution to the problem at hand. In our case, a chromosome is an arrangement, or placement, of all the jigsaw puzzle pieces. Specifically, our GA starts with 1,000 random placements. In every generation the entire population is evaluated using a fitness function (described below), and a new population is (re)produced by the selection of and crossover application to chromosome pairs. The selection method, called {\em roulette wheel selection}, is very common. The probability of selecting a certain chromosome by the method is directly proportionate to the value of its fitness function, as required.

\begin{algorithm}
\caption{Pseudocode of GA framework}
\label{alg:GAMainLoop}
\begin{algorithmic}[1]
\State {$population \gets $ generate 1000 random chromosomes}
\For{$generation\_number = 1 \to 100$}
    \State {evaluate all chromosomes using the fitness function}
    \State $new\_population \gets NULL$
    \State {copy 4 best chromosomes to $new\_population$}
    \While{$size(new\_population) \leq 1000$}
        \State {$parent1 \gets $ select chromosome}
        \State {$parent2 \gets $ select chromosome}
        \State {$child \gets crossover(parent1, parent2)$}
        \State {add child to $new\_population$}
    \EndWhile
    \State {$population \gets new\_population$}
\EndFor
\end{algorithmic}
\end{algorithm}

Having provided a framework overview, we now describe in greater detail the various critical components of the GA proposed, {\eg} the chromosome representation, fitness function, and crossover operator.

\subsection{The fitness function}
The {\em fitness function} (described below) is evaluated for all chromosomes for the purpose of selection. In our GA, each chromosome represents a complete solution to the jigsaw puzzle problem (see Subsection~\ref{sec:repAndCross}), {\ie} a suggested placement of all pieces. The problem variant at hand assumes no knowledge whatsoever of the original image and thus, the correctness of the absolute location of puzzle pieces cannot be estimated in a simple manner. Instead, the pairwise compatibility (defined below) of every pair of adjacent pieces is computed.

We refer to a measure which predicts the likelihood of two pieces to be adjacent in the original image as compatibility. Let $C$ denote this measure. Given two puzzle pieces $x_{i}$, $x_{j}$ and a spatial relation between them $R\in\{l,r,u,d\}$, $C(x_{i}, x_{j}, R)$ denotes the compatibility of piece $x_{j}$ when placed to the left, right, up or down side of piece $x_{i}$, respectively.

Cho {\etal}~\cite{conf/cvpr/ChoAF10} explored five possible compatibility measures, of which the {\em dissimilarity measure} of Eq. (\ref{eq:dissimilarity}) was shown to be the most discriminative. Pomeranz {\etal}~\cite{conf/cvpr/PomeranzSB11} further investigated this issue and chose a similar dissimilarity measure with some slight optimizations. The dissimilarity measure relies on the premise that adjacent jigsaw pieces in the original image tend to share similar colors along their abutting edges, and thus, the sum (over all neighboring pixels) of squared color differences (over all color bands) should be minimal. Assuming pieces $x_{i}$, $x_{j}$ are represented in normalized L*a*b* space by a $K \times K \times 3$ matrix, where $K$ is the height/width of a piece (in pixels), their dissimilarity where $x_{j}$ is to the right of $x_{i}$, for example, is
\begin{equation} \label{eq:dissimilarity}
D(x_{i},x_{j},r)=\sqrt{\sum_{k=1}^{K}\sum_{b=1}^{3}(x_{i}(k,K,b)-x_{j}(k,1,b))^{2}}.
\end{equation}
It is important to note that dissimilarity is not a metric as almost always $D(x_{i},x_{j},R) {\neq} D(x_{j},x_{i},R)$. Obviously, to maximize the compatibility of two pieces, their dissimilarity should be minimized.

Another important consideration in choosing a fitness function is that of run-time cost. Since every chromosome in every generation must be evaluated, a fitness function must be relatively computationally-inexpensive. We chose using the standard dissimilarity, as it meets this criterion and also seems to be sufficiently discriminative. To speed up further the computation of the fitness function we added a lookup table of size $2 \cdot (N \cdot M)^2$ containing all of the pairwise compatibilities for all pieces (we only had to keep compatibilities of the right and up directions since left and down can be easily deduced).

Finally, the fitness function of a given chromosome is the sum of pairwise dissimilarities over all neighboring pieces (whose configuration is represented by the chromosome). Representing a chromosome by an $(N \times M)$ matrix, where a matrix entry $x_{i,j} (i = 1..N, j = 1..M)$ corresponds to a single puzzle piece, we define its fitness as
\begin{equation} \label{eq:fitness}
\sum_{i=1}^{N}\sum_{j=1}^{M-1}(D(x_{i,j},x_{i,j+1},r))+\sum_{i=1}^{N-1}\sum_{j=1}^{M}(D(x_{i,j},x_{i+1,j},d))
\end{equation}
where $r$ and $d$ stand for "right" and "down", respectively.

\subsection{Representation and crossover}
\label{sec:repAndCross}

\begin{figure*}
\centering
        \begin{subfigure}[t]{0.20\textwidth}
                \centering
                \includegraphics[width=\textwidth]{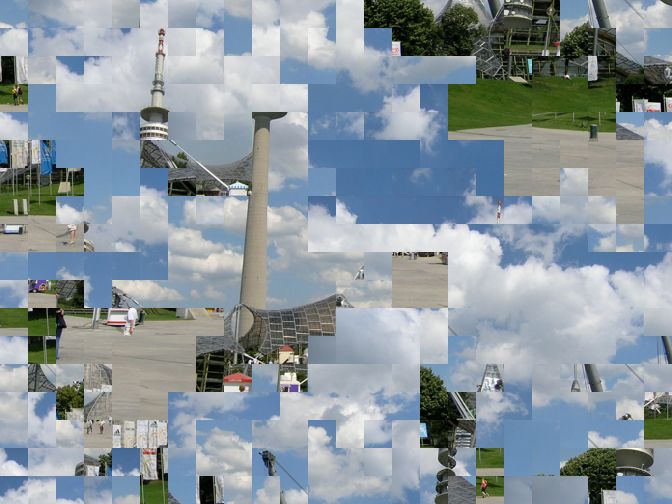}
                \caption{Parent1}
                \label{fig:result_10375_04_gen_00000000}
        \end{subfigure}%
        ~ 
        \begin{subfigure}[t]{0.20\textwidth}
                \centering
                \includegraphics[width=\textwidth]{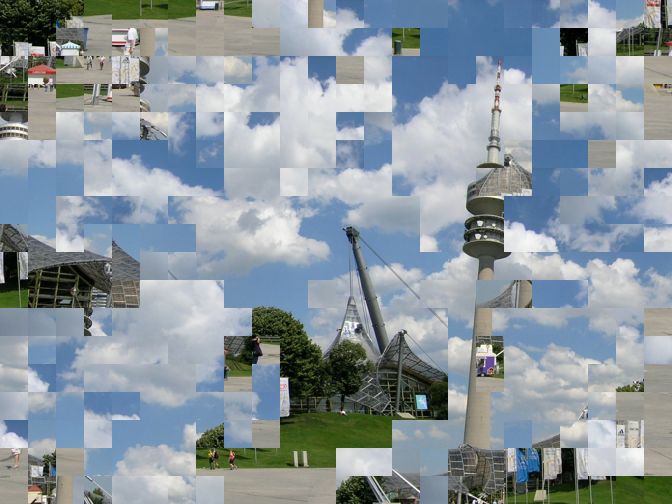}
                \caption{Parent2}
                \label{fig:result_10375_04_gen_00000001}
        \end{subfigure}
        ~ 
        \begin{subfigure}[t]{0.20\textwidth}
                \centering
                \includegraphics[width=\textwidth]{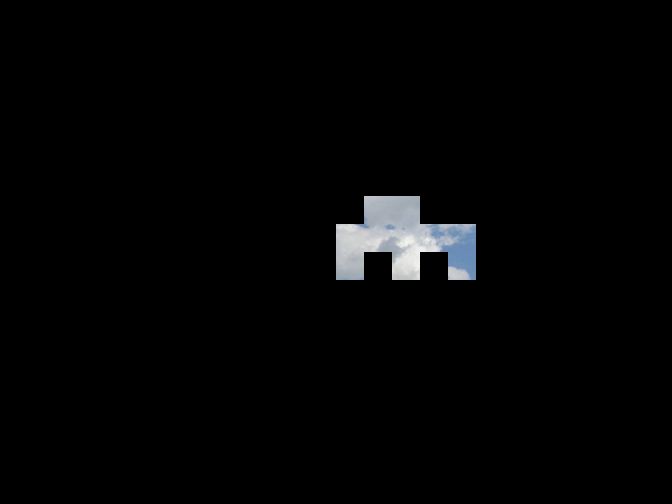}
                \caption{10 Pieces}
                \label{fig:result_10375_04_gen_00000002}
        \end{subfigure}
        ~
        \begin{subfigure}[t]{0.20\textwidth}
                \centering
                \includegraphics[width=\textwidth]{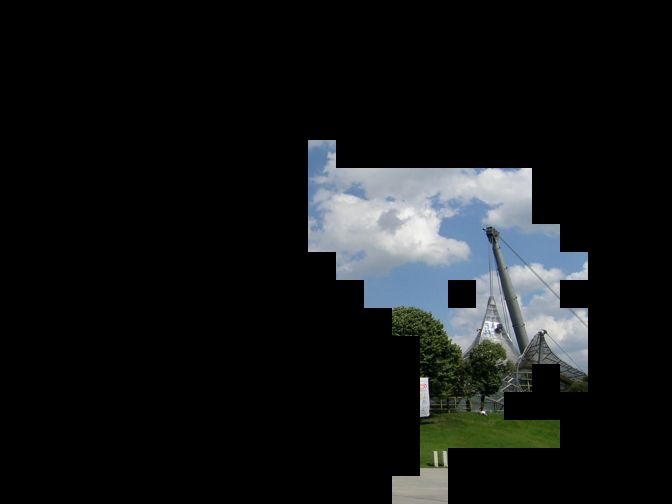}
                \caption{70 Pieces}
                \label{fig:result_10375_04_gen_00000002}
        \end{subfigure}
        ~
        \begin{subfigure}[t]{0.20\textwidth}
                \centering
                \includegraphics[width=\textwidth]{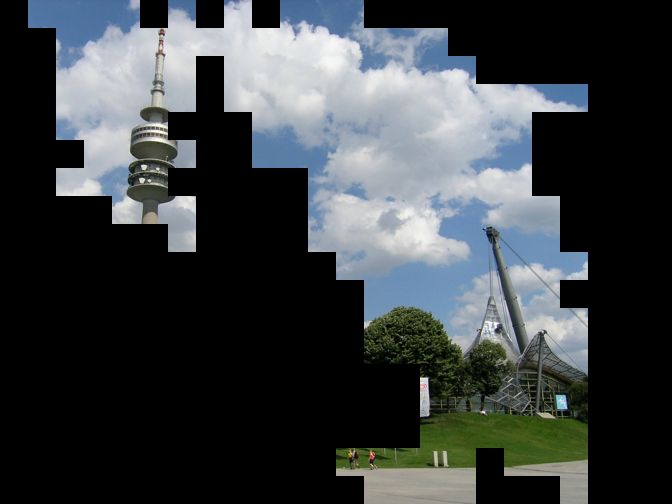}
                \caption{180 Pieces}
                \label{fig:result_10375_04_gen_00000002}
        \end{subfigure}
        ~
        \begin{subfigure}[t]{0.20\textwidth}
                \centering
                \includegraphics[width=\textwidth]{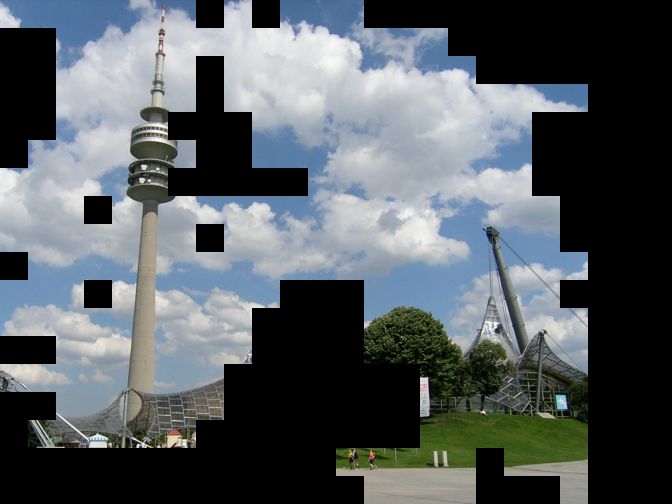}
                \caption{258 Pieces}
                \label{fig:result_10375_04_gen_00000002}
        \end{subfigure}
        ~
        \begin{subfigure}[t]{0.20\textwidth}
                \centering
                \includegraphics[width=\textwidth]{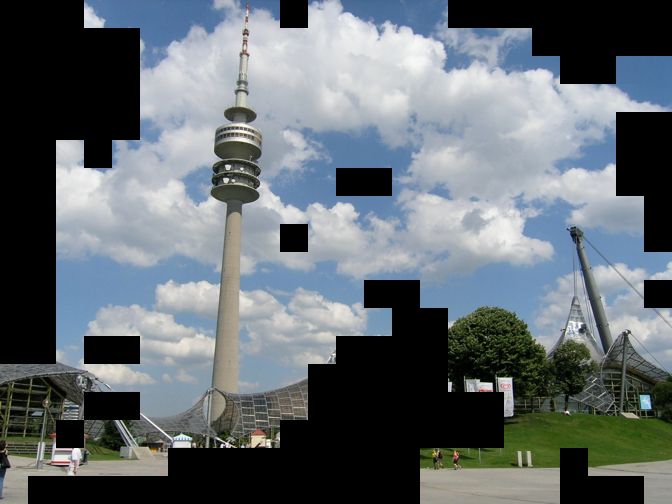}
                \caption{304 Pieces}
                \label{fig:result_10375_04_gen_00000002}
        \end{subfigure}
        ~
        \begin{subfigure}[t]{0.20\textwidth}
                \centering
                \includegraphics[width=\textwidth]{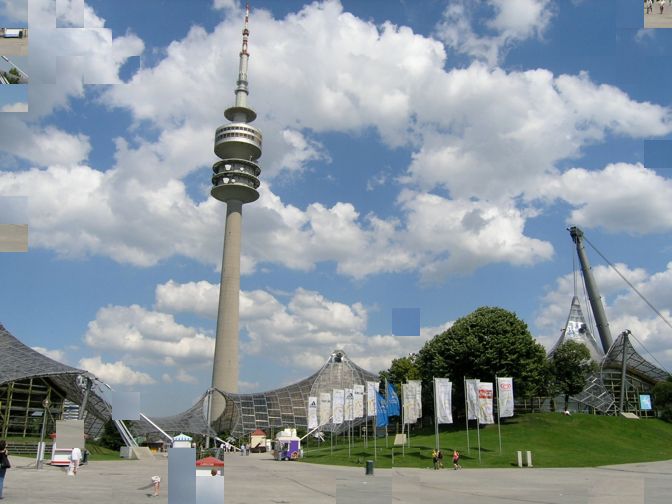}
                \caption{Child}
                \label{fig:result_10375_04_gen_00000100}
        \end{subfigure}
        \caption{ Illustration of crossover operation: Given (a) Parent1 and (b) Parent2, (c) -- (g) depict how a kernel of pieces is gradually grown until (h) a complete child. Note the detection of parts of the tower in both parents, which are then shifted and merged to the complete tower; shifting of images during kernel growing is due to piece position independence. }
        \label{fig:growingKernel}
\end{figure*}

\subsubsection{Problem definition}

As noted above, given a puzzle (image) of $(N \times M)$ pieces, a chromosome may be represented by an $(N \times M)$ matrix, each entry of which corresponds to a piece number. (A piece is assigned a number according to its initial location in the given puzzle.) This representation is straightforward and lends itself easily to the evaluation of the fitness function described above. The main issue stemming from this representation is the design of an appropriate crossover operator. As previously noted, this operator receives two parent chromosomes and creates a child chromosome. It should allow for "good traits" from the parents to pass on to the child, thereby creating possibly a better solution. A naive crossover operator with respect to the given representation will create a new child chromosome at random, such that each entry of the resulting matrix is the corresponding cell of the first or second parent. This approach yields usually a child chromosome with duplicate and/or missing puzzle pieces, which makes of course an invalid solution to the problem. It seems that the inherent difficulty surrounding the crossover issue may have played a critical role in delaying thus far the development of a state-of-the-art GA solution to the problem.

Once the validity issue is rectified, one still needs to consider very carefully the crossover operator. Recall, crossover is applied to two chromosomes selected due to their high fitness values, where the fitness function used is an overall pairwise compatibility measure of adjacent puzzle pieces. At best, the function rewards a correct placement of neighboring pieces next to each other, but it has no way of identifying the correct absolute location of a piece. Since the population starts out from a random piece placement and then gradually improves, it is reasonable to assume that over the generations some correctly assembled puzzle segments will emerge. Taking into account the fitness function's inability to reward a correct position, we expect such segments to appear most likely at incorrect absolute locations. Discovering a correct segment is not trivial; it should be regarded a good trait that needs to be exploited by passing it on to the child chromosome. The crossover operator must allow for position independence, {\ie} the ability of shifting correct segments, so as to place them correctly ({\ie} in their correct absolute location) in the child.

Finally, once the position-independence issue is settled, one should address the issue of detecting these aforementioned, possibly misplaced, correct segments. What segment should the crossover operator pass on to an offspring? A random approach might seem appealing, but in practice it could be infeasible due to the enormous size of the problem's solution domain. Some heuristics may be applied to distinguish correct segments from incorrect ones.

\subsubsection{Our proposed solution}

Given two parent chromosomes, {\ie} two complete (different) arrangements of all puzzle pieces, the crossover operator constructs a child chromosome in a kernel-growing fashion, using both parents as "consultants". The operator starts with a single piece and gradually joins other pieces at available boundaries. New pieces may be joined only adjacently to existing pieces, so that the emerging image is always contiguous. The operator keeps adding pieces from a bank of available pieces until there are no more pieces left. Hence, every piece will appear exactly once in the resulting image. Since the image size is known in advance, the operator can ensure no boundary violation. Thus, by using every piece exactly once inside of a frame of the correct size, the operator is guaranteed of achieving a valid image. Figure~\ref{fig:growingKernel} illustrates the above kernel-growing process.

A key trait of the kernel-growing technique is the fact that the final absolute location of every piece is determined only once the kernel reaches its final size and the child chromosome is complete. Until that point, all pieces might be shifted, depending the kernel's growth vector. The first piece, for example, might eventually be located at the lower-left corner of the image, should the kernel grow only to the up and to the right, after this piece was assigned. Instead, the same first piece might ultimately be located at the center of the image, upper-right corner, or any other location. This change in the absolute location of each piece is illustrated in Figure~\ref{fig:growingKernel}, especially between phase (f) and phase (g) of the kernel-growing process, as all pieces are shifted to the right due to insertion of new pieces on the left. It is this important trait which enables the position independence of image segments. 

Now remains the question of which piece to select from the available pieces bank and where to locate it in the child. Given a kernel, {\ie} a partial image, we can mark all the boundaries where a new piece might be placed. A piece boundary is denoted by a pair $(x_{i}, R)$, consisting of the piece number and a spatial relation. The operator invokes a three-phase procedure. First, given all existing boundaries, the operator checks whether there exists a piece boundary for which both parents agree on a piece $x_{j}$ (meaning, both contain this piece in the spatial direction $R$ of  $x_{i}$). If such a piece exists, then it is placed in the correct location. If the parents agree on two or more boundaries, one of them is chosen at random and the respective piece is assigned. Obviously, an already used piece cannot be (re)assigned, so any such piece is ignored as if the parents did not agree on that particular boundary. If there is no agreement between the parents on any piece at any boundary, the second phase begins. To understand this phase, we briefly review the concept of a {\em best-buddy} piece, first introduced by Pomeranz {\etal}~\cite{conf/cvpr/PomeranzSB11}; two pieces are said to be best-buddies if each piece considers the other as its most compatible piece. The pieces $x_{i}$ and $x_{j}$ are said to best-buddies if
\begin{align}
\forall x_{k} \in Pieces, \; C(x_{i},x_{j},R_1) \geq C(x_{i},x_{k},R_1)\notag \\
\text{and \quad\quad\quad\quad\quad\quad\quad\quad}\\
\forall x_{p} \in Pieces, \; C(x_{j},x_{i},R_2) \geq C(x_{j},x_{p},R_2) \notag
\end{align}
where $Pieces$ is the set of all given image pieces and $R_1$ and $R_2$ are "complementary" spatial relations ({\eg} if $R_1$ = right, then $R_2$ = left and vice versa). In the second phase the operator checks whether one of the parents contains a piece $x_j$ in spatial relation $R$ of $x_i$ which is also a best-buddy of $x_i$ in that relation. If so, the piece is chosen and assigned. As before, if multiple best-buddy pieces are available, one is chosen at random. If a best-buddy piece is found but was already assigned, it is ignored and the search continues for other best-buddy pieces. Finally, if no best-buddy piece exists, the operator randomly selects a boundary and assigns it the most compatible piece available. To introduce mutation -- in the first and last phase the operator places, with low probability, an available piece at random, instead of the most compatible relevant piece available. 

In summary, the operator uses repeatedly a three-phase procedure of piece selection and assignment, placing first agreed pieces, followed by best-buddy pieces and finally by the most compatible piece available ({\ie} not already assigned). An assignment is only considered at relevant boundaries to maintain the contiguity of the kernel-growing image. The procedure returns to the first phase after every piece assignment due to the prospective creation of new boundaries. A simplified description of the crossover operator (without mutation) can be found in Algorithm~\ref{alg:CrossoverSimplified}.

\begin{algorithm}
\caption{Crossover operator simplified}
\label{alg:CrossoverSimplified}
\begin{algorithmic}[1]
\State {If any available boundary meets the criterion of Phase 1 (both parents agree on a piece), place the piece there and goto (1); otherwise continue.}
\State {If any available boundary meets the criterion of Phase 2 (one parent contains a best-buddy piece), place the piece there and goto (1); otherwise continue.}
\State {Randomly choose a boundary, place the most compatible available piece there and goto (1).}
\end{algorithmic}
\end{algorithm}

\subsubsection{Rationale}

In a GA framework, good traits should be passed on to the child. Here, since position independence of pieces is encouraged, the trait of interest is captured by a piece's set of neighbors. Correct puzzle segments correspond to a correct placement of pieces next to each other. The notion that piece $x_{i}$ is in spatial relation $R$ to piece $x_{j}$  is key to solving the jigsaw problem. Nevertheless, every chromosome accounts for a complete placement of all the pieces. Taking into account the random nature of the first generation, the procedure must actively seek the better traits of piece relations. In our work, we assume that a trait common to both parents has propagated through the generations and comprises the reason for their survival and selection. In other words, if both parents agree on a relation, we regard it as true with high probability. Note that not all agreed relations are copied immediately to the child. Since a kernel-growing algorithm is used, some agreed pieces might "prematurely" serve as most compatible pieces at another boundary and be subsequently disqualified for later use. Thus, random agreements in early generations are likely to be nullified.

As for the second stage, where the parents agree on no piece, one might be inclined to randomly pick a parent and follow its lead. Another option might be to just choose the most compatible piece in a greedy manner, or check  if a best-buddy piece is available. Since piece placement in parents might be random and since even best-buddy pieces might not capture the correct match, we combine the two. The fact that two pieces are both best-buddies and are adjacent at a parent is a good indication for the validity of this match. A different perspective is to consider that every chromosome contains some correct segments. The passing of correct segments from parents to children is at the heart of the GA. Moreover, if each parent contains a correct segment and these segments partially overlap, the overlapping (agreed upon) part will be copied to the child in the first phase and be completed from both parents at the second stage, thus combining the segments into a larger correct segment, creating a better child solution, and advancing the pursuit of the entire correct image.

As for the more greedy third step, we may conclude that the GA concurrently tries many different greedy placements, and only those that seem correct propagate through the generations. This exemplifies the principle of propagation of good traits in the spirit of the theory of natural selection.

\section{Experimental results}

\begin{figure*}
\centering
        \begin{subfigure}[t]{0.20\textwidth}
                \centering
                \includegraphics[width=\textwidth]{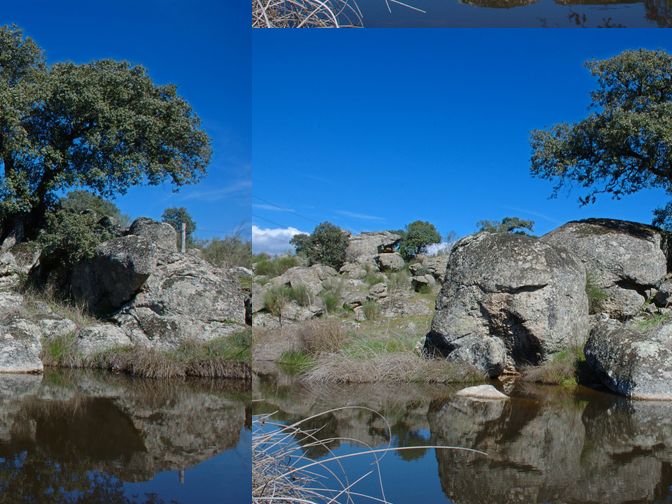}
                \caption{}
                \label{fig:result_10375_04_gen_00000000}
        \end{subfigure}%
        ~ 
        \begin{subfigure}[t]{0.20\textwidth}
                \centering
                \includegraphics[width=\textwidth]{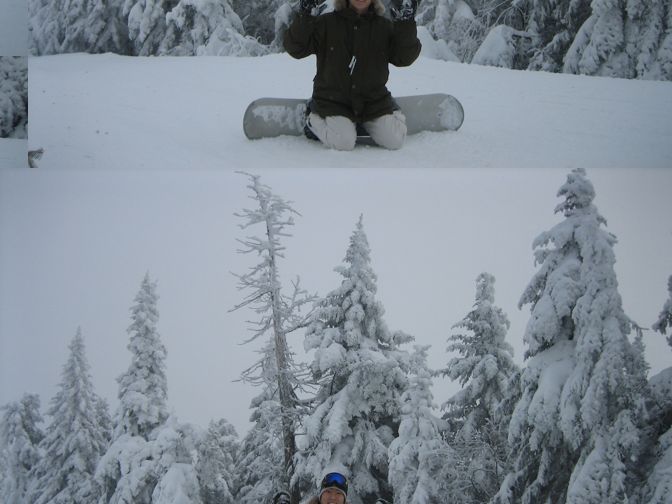}
                \caption{}
                \label{fig:result_10375_04_gen_00000001}
        \end{subfigure}
        ~ 
        \begin{subfigure}[t]{0.20\textwidth}
                \centering
                \includegraphics[width=\textwidth]{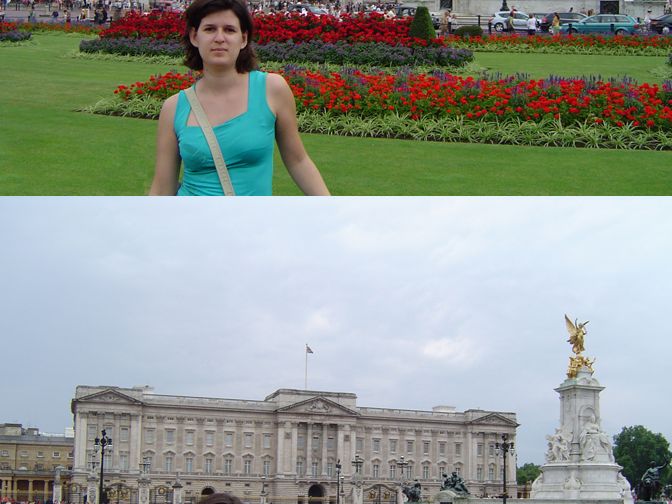}
                \caption{}
                \label{fig:result_10375_04_gen_00000002}
        \end{subfigure}
        ~
        \begin{subfigure}[t]{0.20\textwidth}
                \centering
                \includegraphics[width=\textwidth]{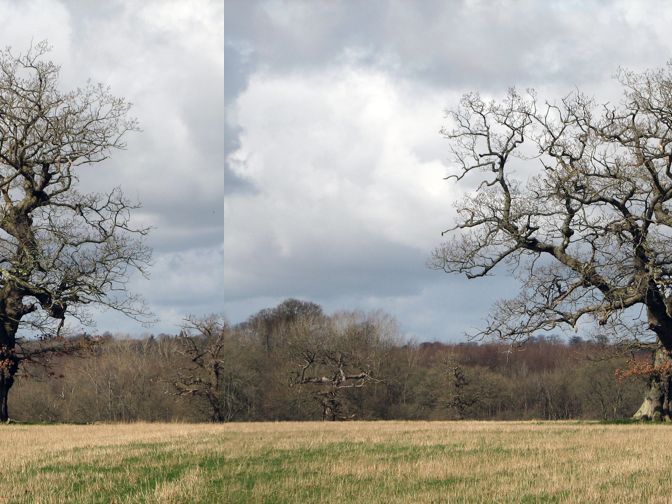}
                \caption{}
                \label{fig:result_10375_04_gen_00000100}
        \end{subfigure}
        \caption{Shifted puzzle solutions. All images are solutions created by our GA. The accuracy for each solution is 0\% according to the direct comparison, but over 95\% according to the (more reasonable) neighbor comparison. Amazingly, the dissimilarity of each solution is smaller than that of its original image counterpart.  }
        \label{fig:shited}
\end{figure*}

Cho {\etal}~\cite{conf/cvpr/ChoAF10} introduced three measures to evaluate the correctness of an assembled puzzle, two of which were repeatedly used in previous works: The {\em direct} comparison which measures the fraction of pieces located in their correct absolute location, and the {\em neighbor} comparison, which measures the fraction of correct neighbors. The direct method has been repeatedly denounced~\cite{conf/cvpr/PomeranzSB11} as being both less accurate and less meaningful due to its inability to cope with slightly shifted puzzle solutions. Figure~\ref{fig:shited} illustrates the drawbacks of the direct comparison and the superiority of the neighbor comparison. Note that a piece arrangement scoring 100\% according to one of the methods is, by definition, the full reconstruction of the original image and will also achieve a score of 100\% when measured by the other method. Thus, unless stated otherwise, all results are under neighbor comparison. For the sake of completeness, our results under direct comparison are reported in Table~\ref{tab:directAvg}.

In all experiments, we used the same GA parameters described in Algorithm~\ref{alg:GAMainLoop}. The population consists of 1000 chromosomes. In each generation we retain the best 4 chromosomes (a measure called {\em elitism}). The rest of the population is generated by the crossover operator described earlier with a mutation rate of 5\%. Parent chromosomes are chosen by the roulette wheel selection, producing a single offspring in each crossover. The GA always runs for exactly 100 generations.

We ran the proposed GA on the set of images supplied by Cho {\etal}~\cite{conf/cvpr/ChoAF10} and all sets supplied by Pomeranz {\etal}~\cite{conf/cvpr/PomeranzSB11}, testing puzzles of 28 $\times$ 28-pixel patches according to the traditional convention. The image data experimented with contains 20-image sets of 432-, 504-, and 805-piece puzzles and 3-image sets of 2,360- and 3,360-piece puzzles. We ran the GA 10 times on each image, each time with a different random seed, and recorded -- over these 10 runs -- the best, worst, and average accuracy (as well as the standard deviation). Table~\ref{tab:bestWorstAvgComp} lists the results achieved by our GA on each set. Interestingly, despite the expected random nature of GAs, the results of different runs were almost identical, attesting to the robustness of the GA. Table~\ref{tab:pomAvgCompare} compares -- for each image set -- our average best results to those of Pomeranz {\etal}, which can be easily derived from their detailed documented results~\cite{conf/cvpr/site/PomeranzSB11}. As can be seen, our GA results are far more accurate than the state-of-the-art. Nevertheless, as noted in the beginning of this paper, previous solvers do well on some images and not as well on others. The superior performance of our solver is best conveyed in Table~\ref{tab:pomAvgWorseCompare}, where we only relate to the 3 least accurately solved images in every set. Our solver gains a significant improvement of up to 21\% (for the 540-piece puzzle set); for some puzzles, the improvement was even 30\%. Detailed results of the exact accuracy of every run on every image can be found in our supplementary material~\cite{conf/cvpr/site/Our}.

\begin{figure*}
\centering
      \begin{subfigure}[t]{0.20\textwidth}
                \centering
                \includegraphics[width=\textwidth]{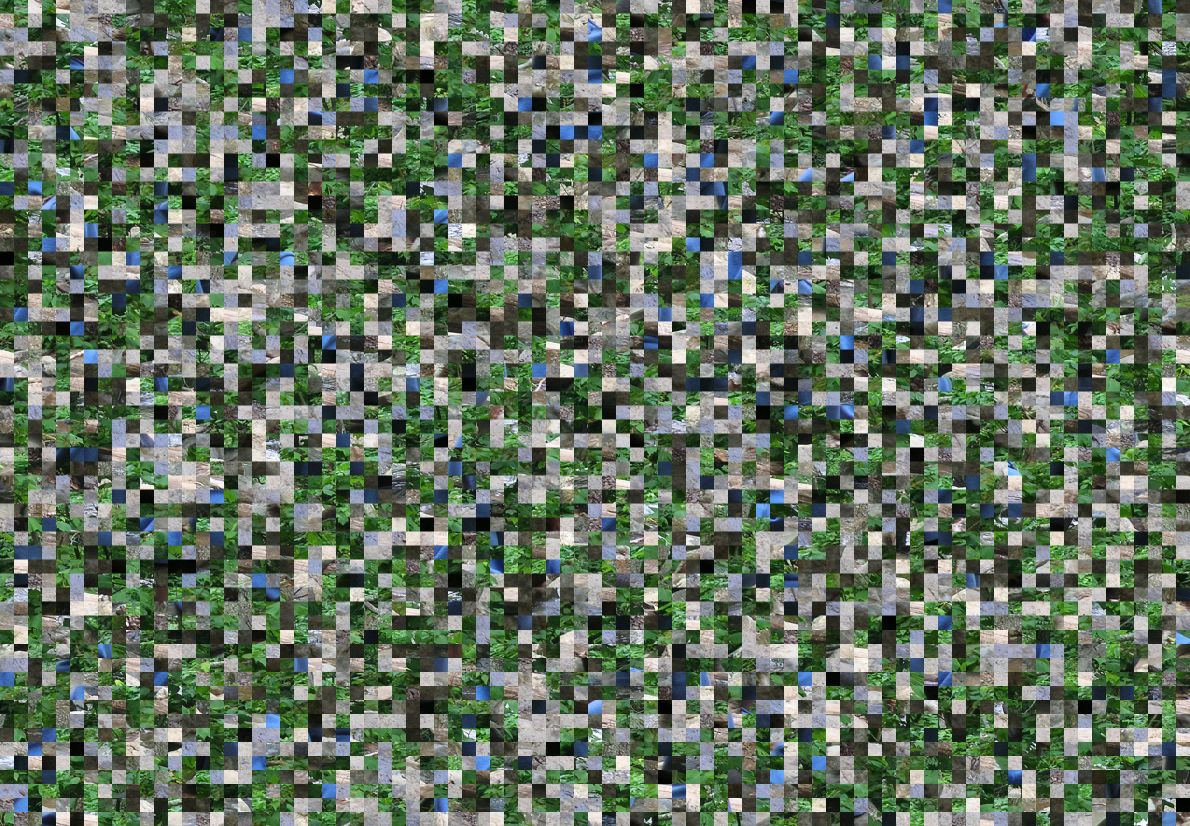}
                \caption{5,015 pieces}
                \label{fig:result_5015_19_gen_00000000}
        \end{subfigure}%
        ~ 
        \begin{subfigure}[t]{0.20\textwidth}
                \centering
                \includegraphics[width=\textwidth]{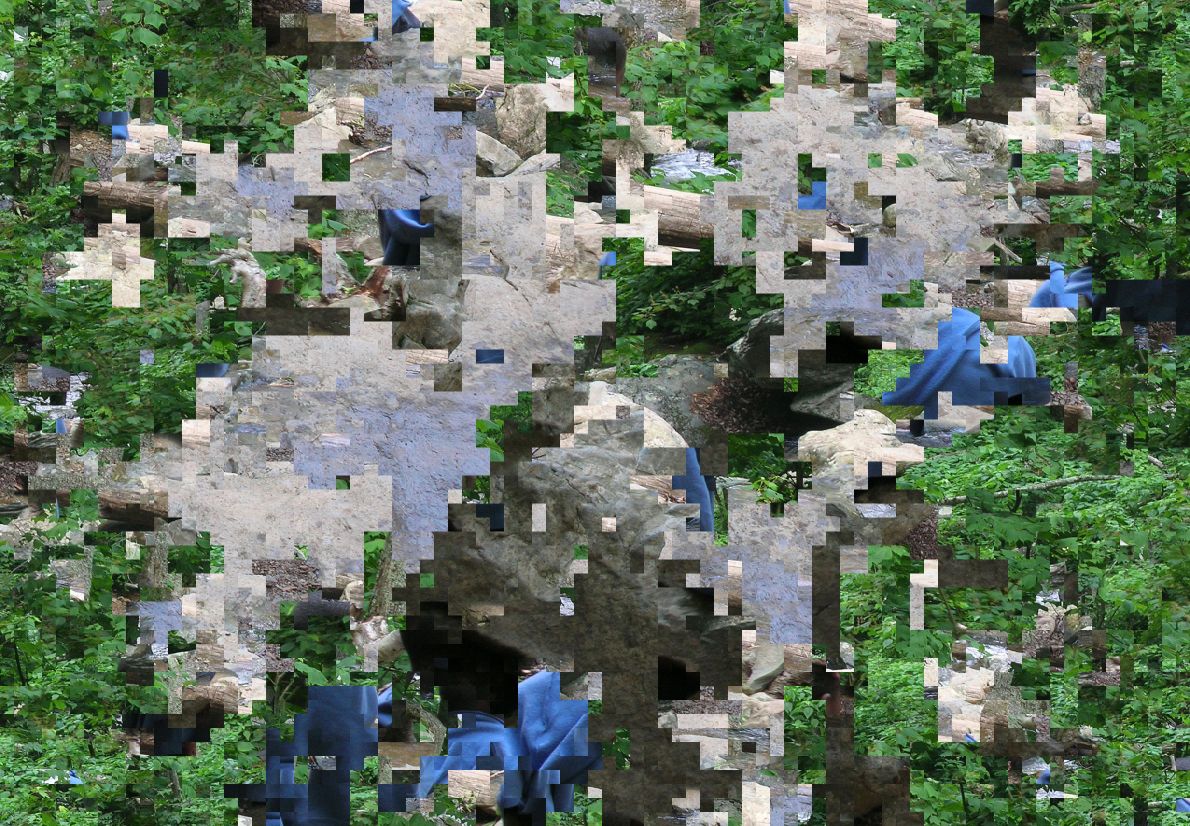}
                \caption{Generation 1}
                \label{fig:result_5015_19_gen_00000001}
        \end{subfigure}
        ~ 
        \begin{subfigure}[t]{0.20\textwidth}
                \centering
                \includegraphics[width=\textwidth]{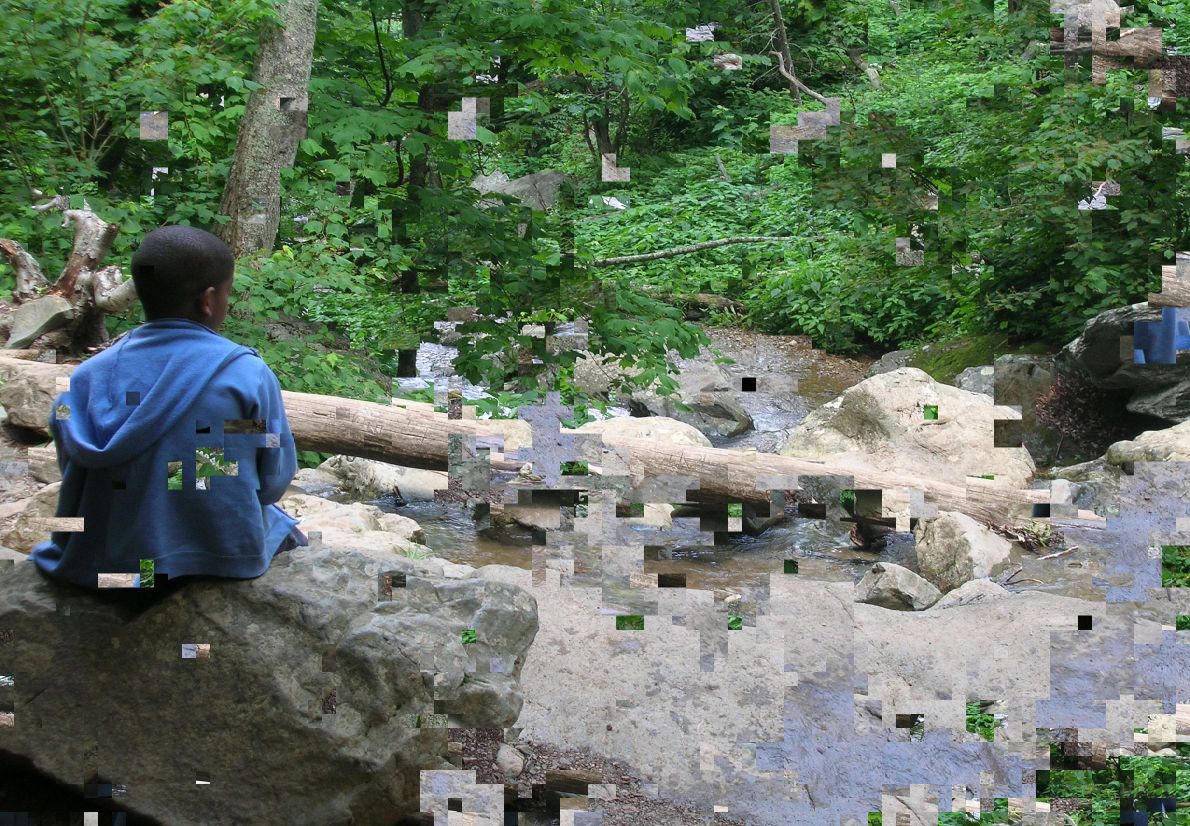}
                \caption{Generation 2}
                \label{fig:result_5015_19_gen_00000002}
        \end{subfigure}
        ~
        \begin{subfigure}[t]{0.20\textwidth}
                \centering
                \includegraphics[width=\textwidth]{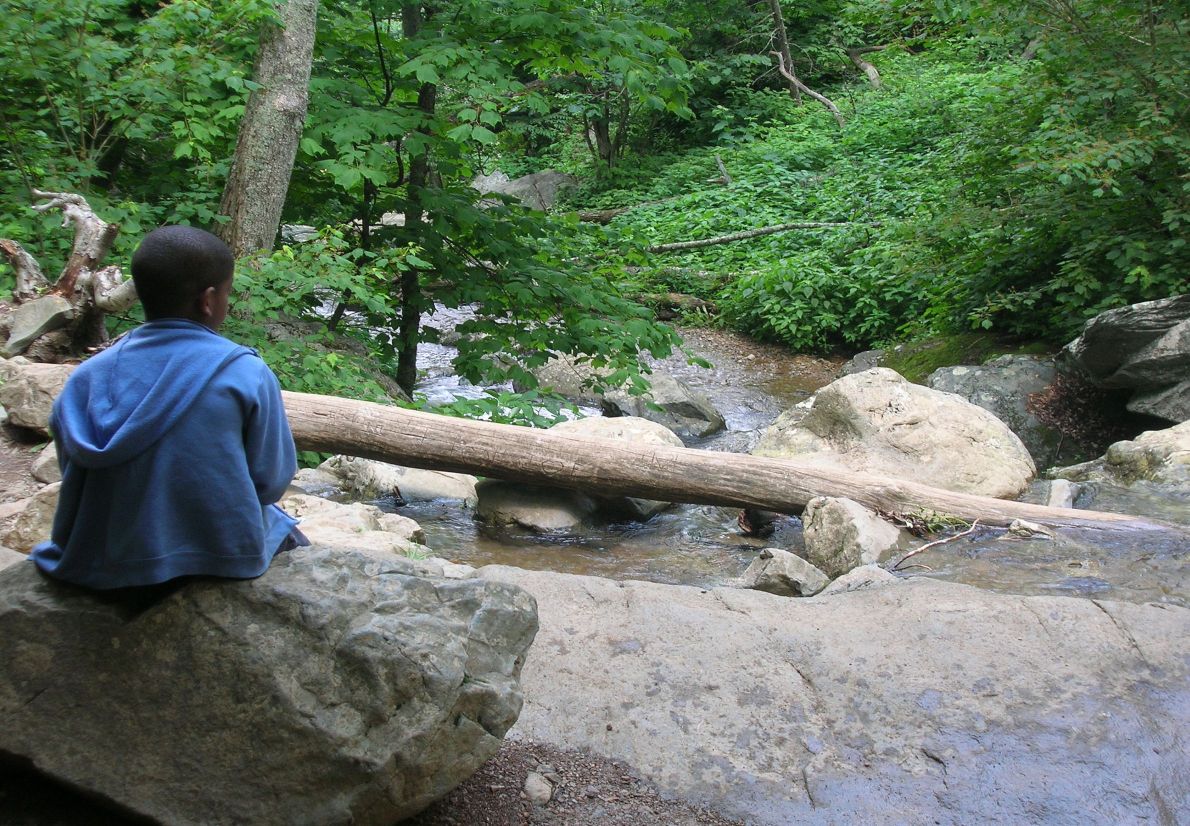}
                \caption{Final}
                \label{fig:result_5015_19_gen_00000100}
        \end{subfigure}

        \begin{subfigure}[t]{0.20\textwidth}
                \centering
                \includegraphics[width=\textwidth]{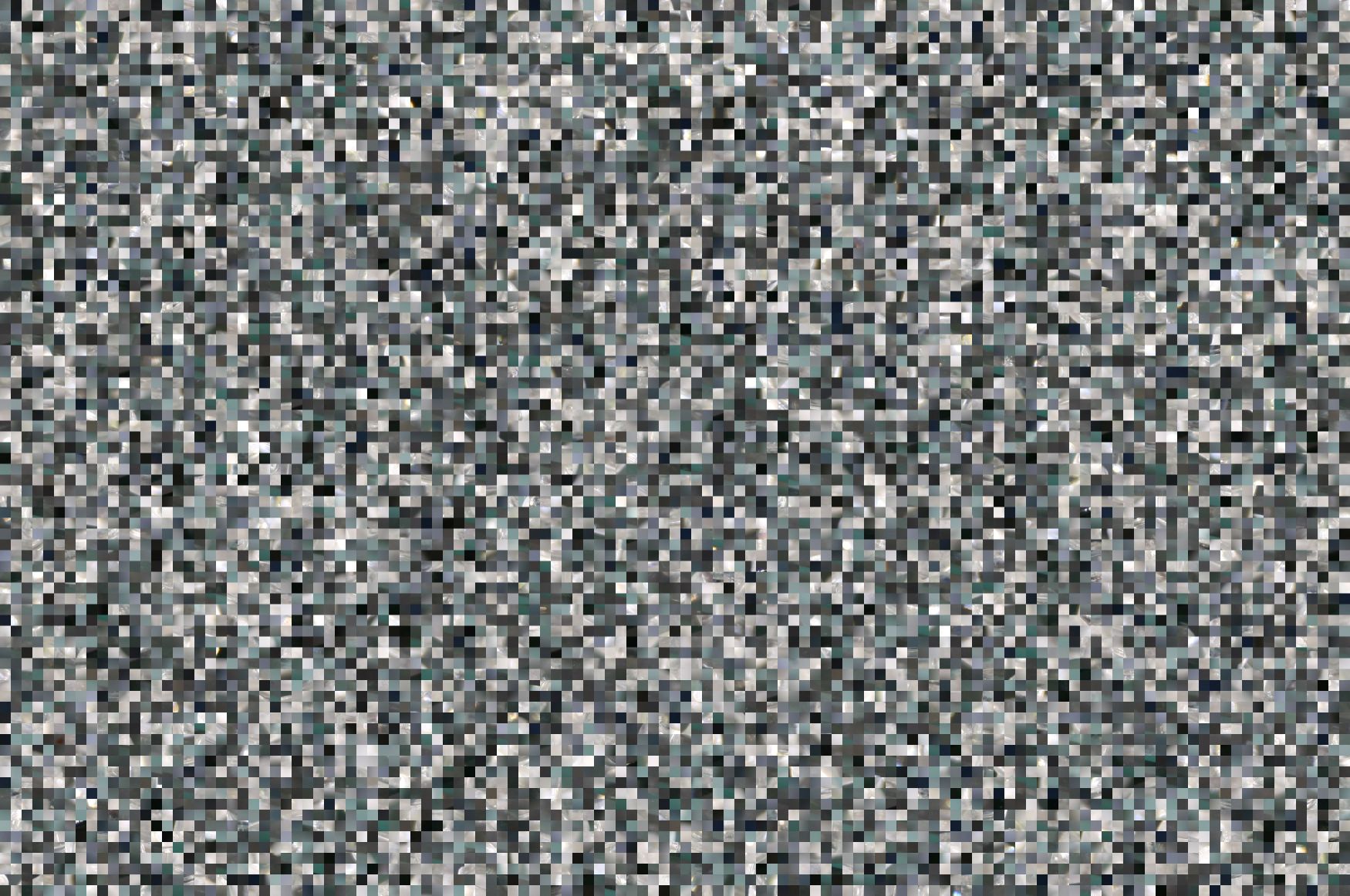}
                \caption{10,375 pieces}
                \label{fig:result_10375_04_gen_00000000}
        \end{subfigure}%
        ~ 
        \begin{subfigure}[t]{0.20\textwidth}
                \centering
                \includegraphics[width=\textwidth]{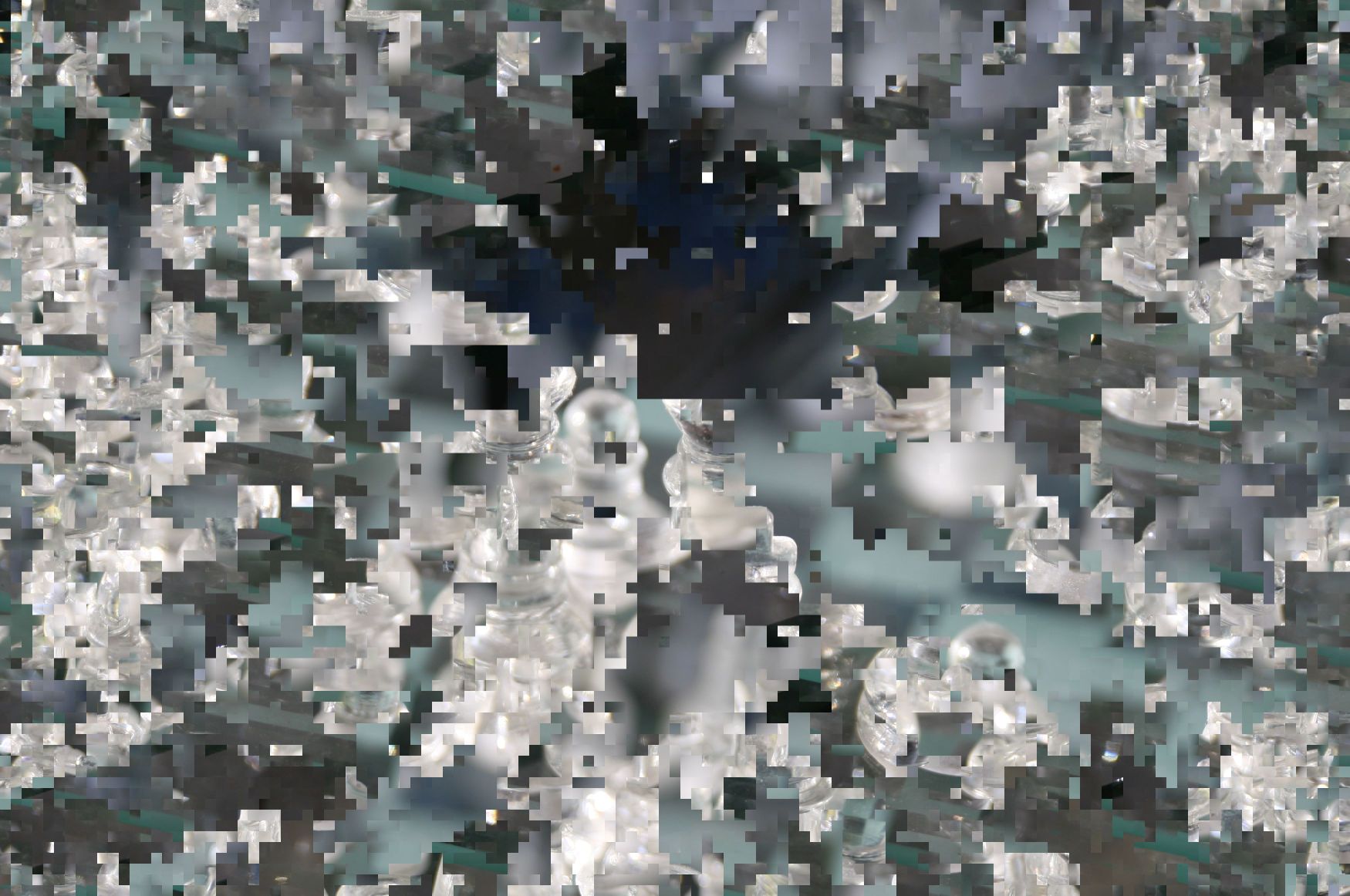}
                \caption{Generation 1}
                \label{fig:result_10375_04_gen_00000001}
        \end{subfigure}
        ~ 
        \begin{subfigure}[t]{0.20\textwidth}
                \centering
                \includegraphics[width=\textwidth]{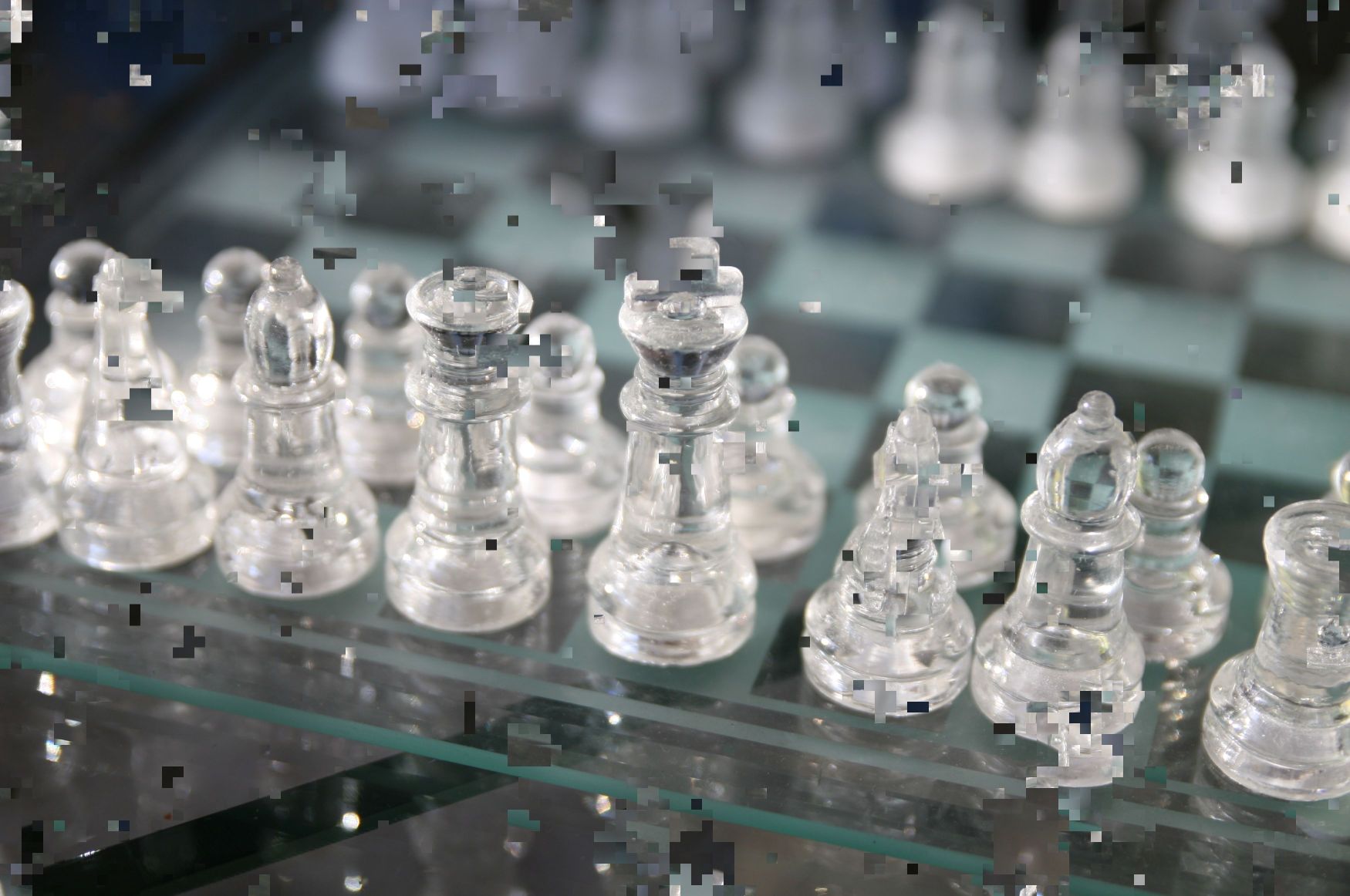}
                \caption{Generation 2}
                \label{fig:result_10375_04_gen_00000002}
        \end{subfigure}
        ~
        \begin{subfigure}[t]{0.20\textwidth}
                \centering
                \includegraphics[width=\textwidth]{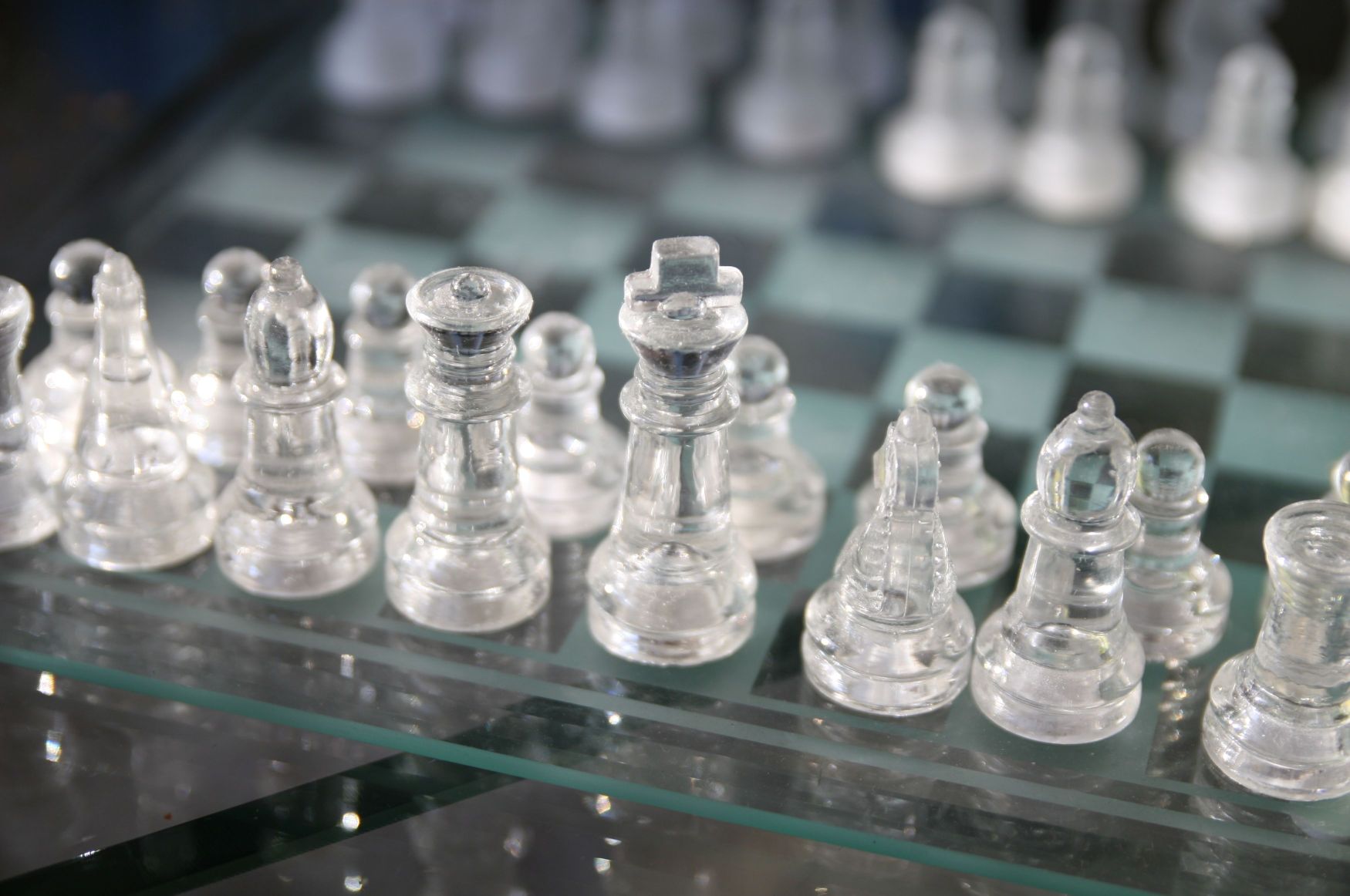}
                \caption{Final}
                \label{fig:result_10375_04_gen_00000100}
        \end{subfigure}

        \begin{subfigure}[t]{0.20\textwidth}
                \centering
                \includegraphics[width=\textwidth]{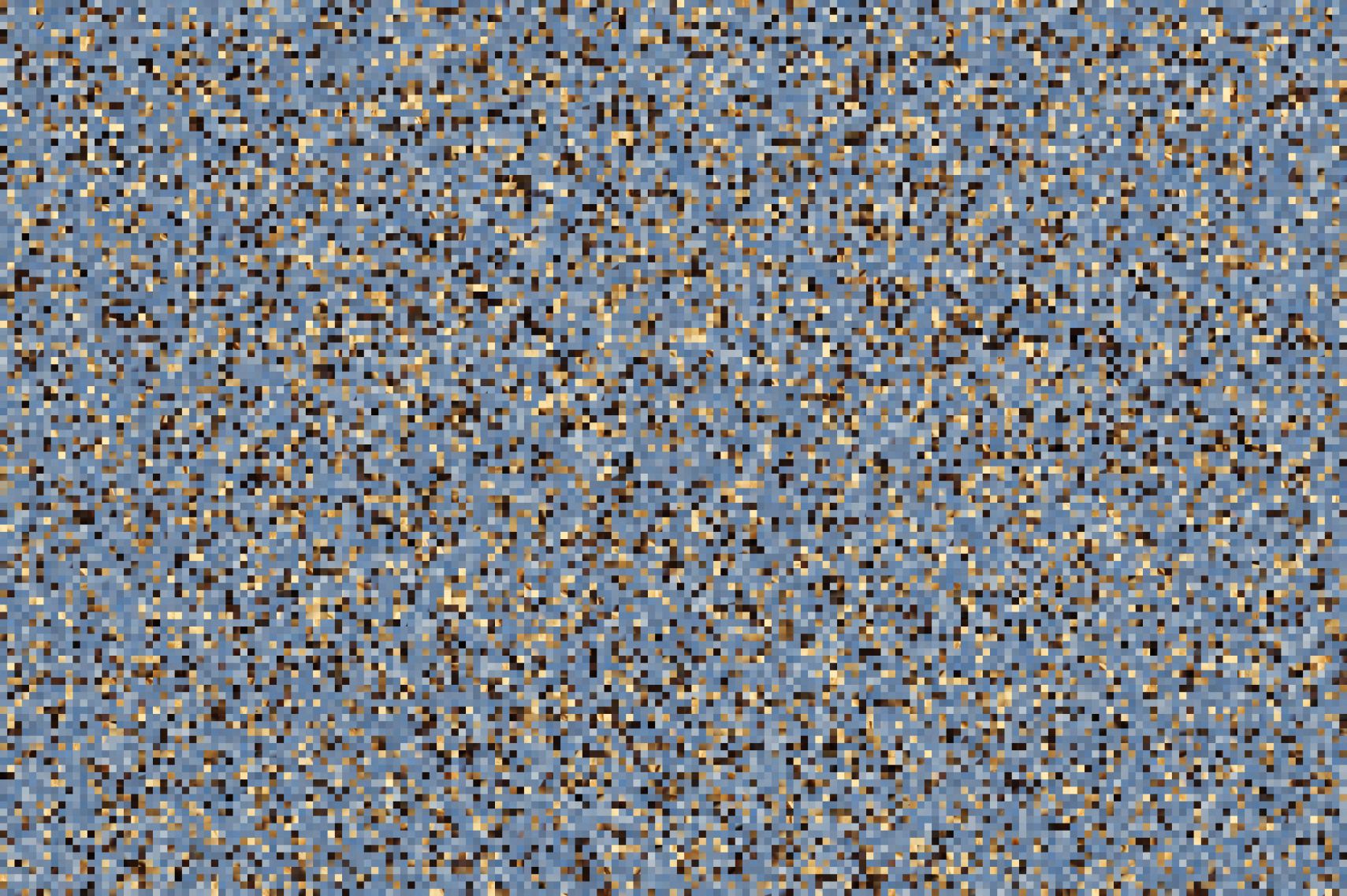}
                \caption{22,834 pieces}
                \label{fig:result_22834_12_gen_00000000}
        \end{subfigure}%
        ~ 
        \begin{subfigure}[t]{0.20\textwidth}
                \centering
                \includegraphics[width=\textwidth]{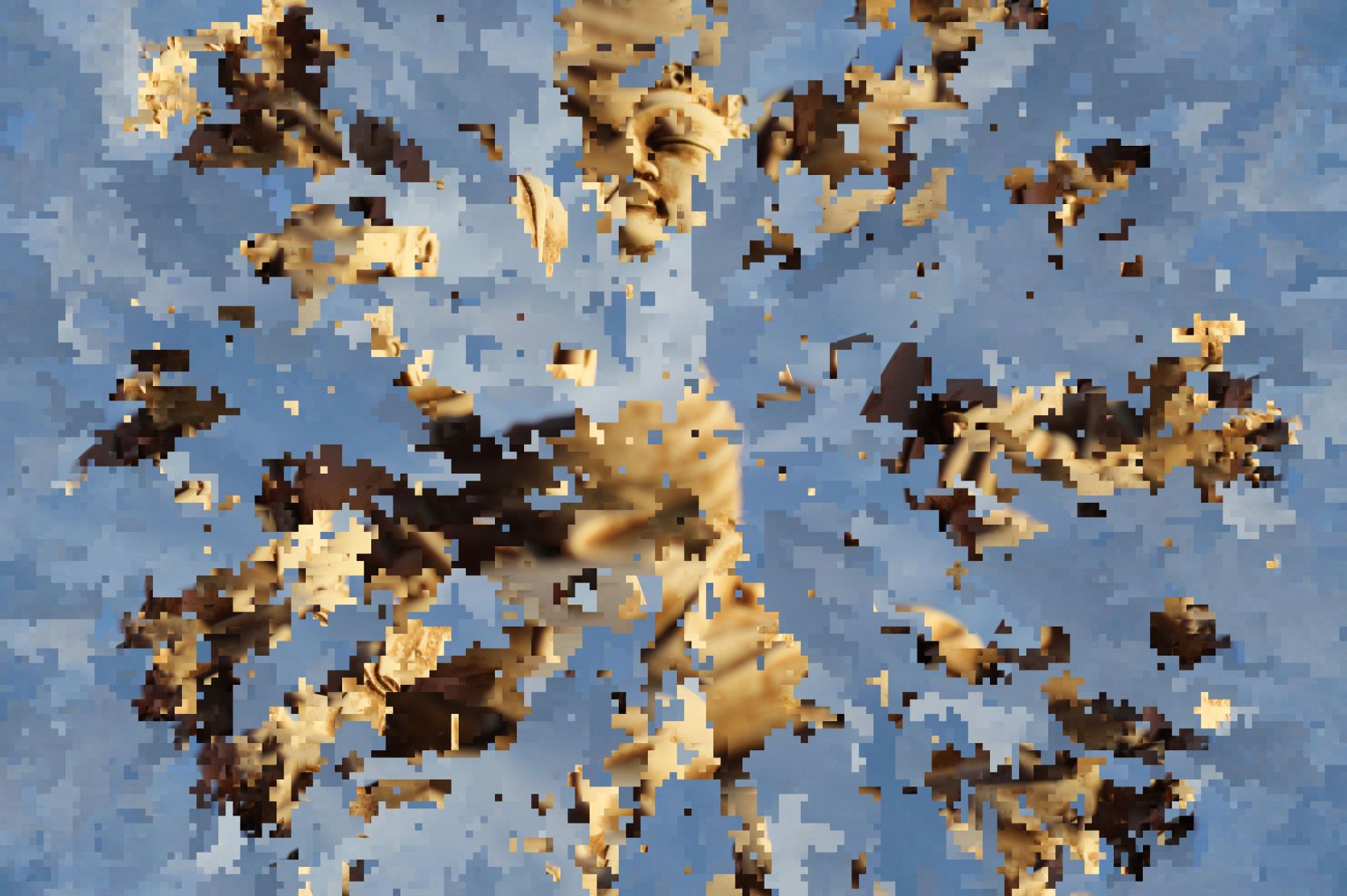}
                \caption{Generation 1}
                \label{fig:result_22834_12_gen_00000001}
        \end{subfigure}
        ~ 
        \begin{subfigure}[t]{0.20\textwidth}
                \centering
                \includegraphics[width=\textwidth]{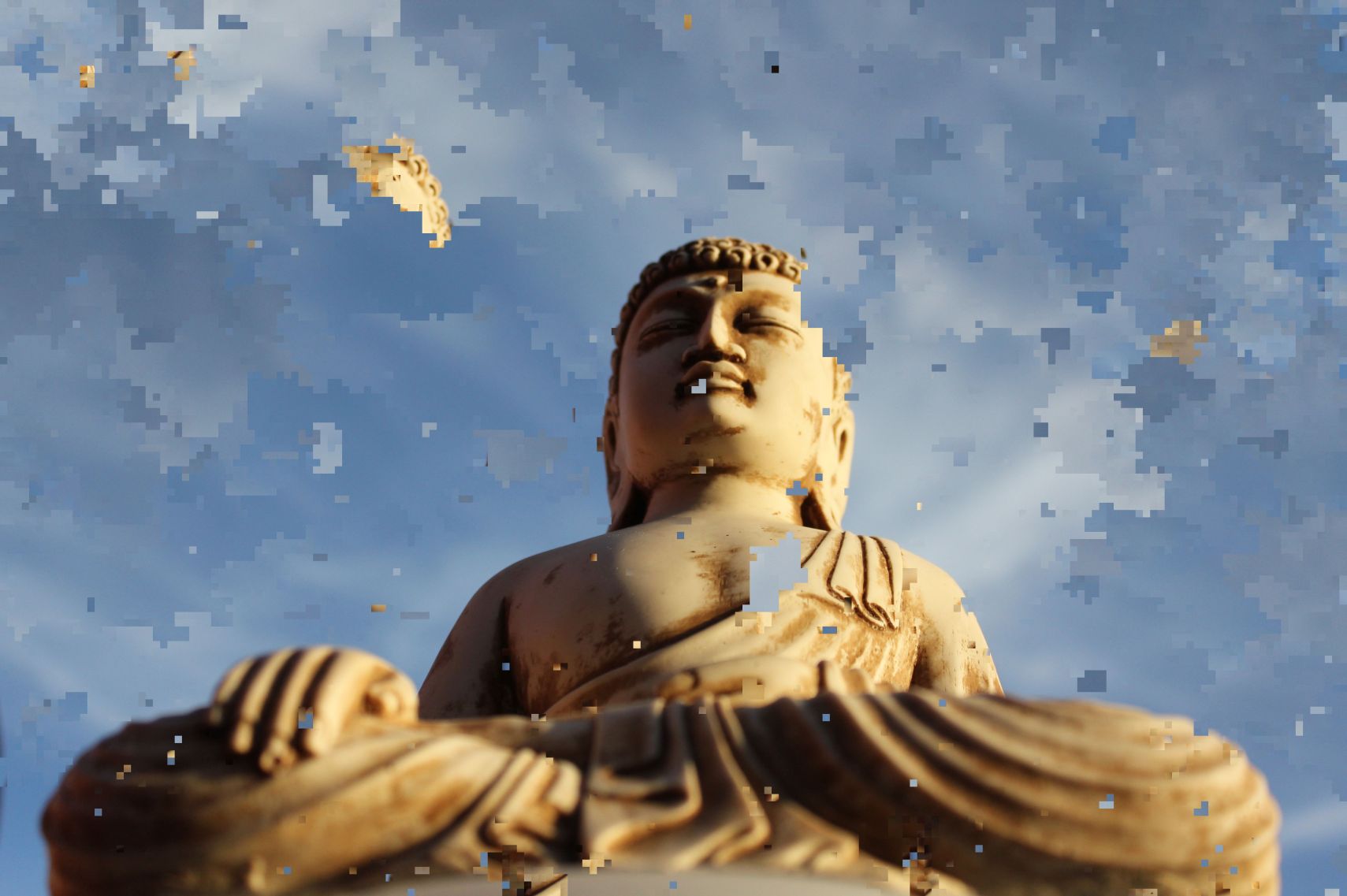}
                \caption{Generation 2}
                \label{fig:result_22834_12_gen_00000002}
        \end{subfigure}
        ~
        \begin{subfigure}[t]{0.20\textwidth}
                \centering
                \includegraphics[width=\textwidth]{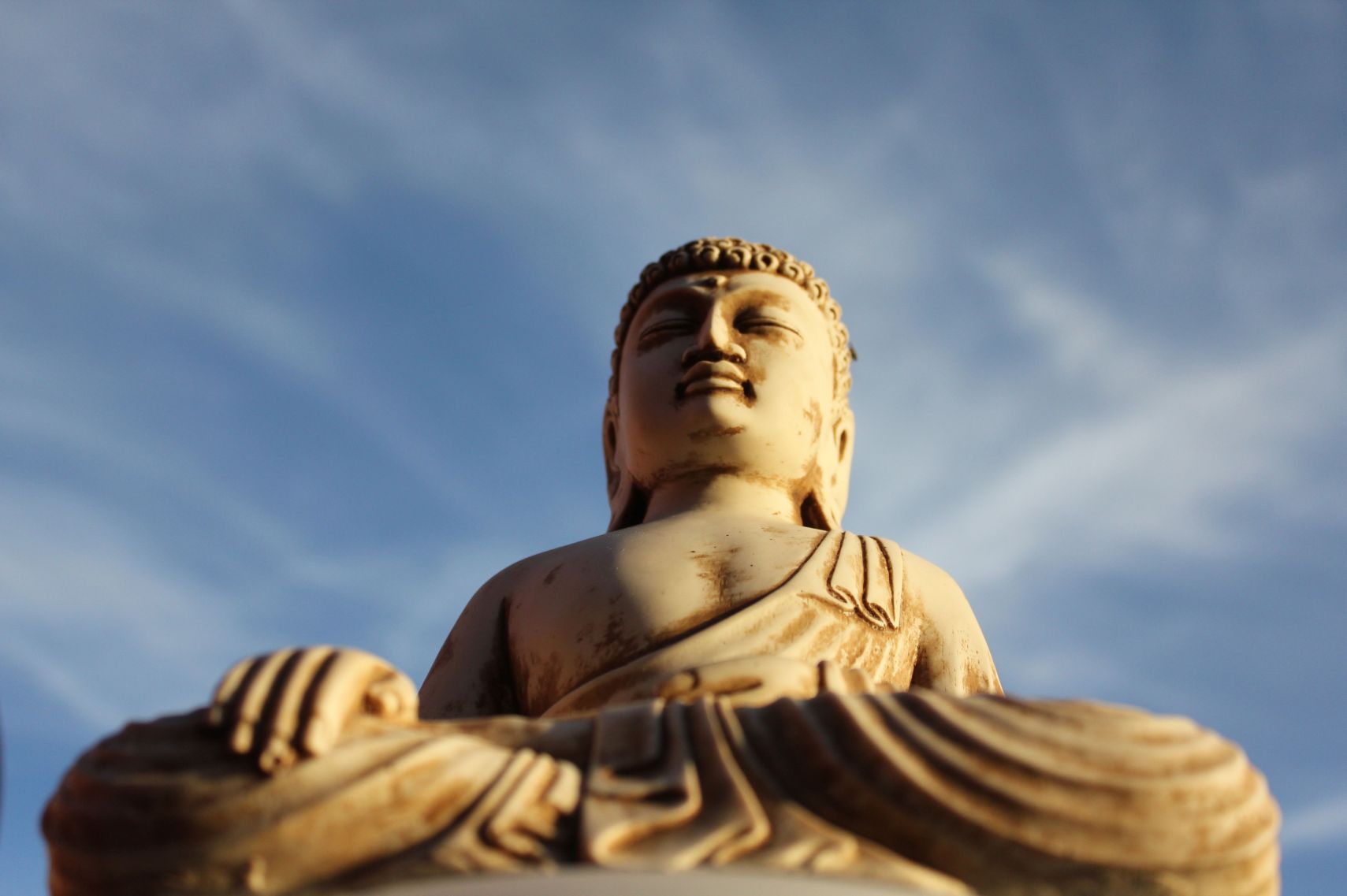}
                \caption{Final}
                \label{fig:result_22834_12_gen_00000007_final}
        \end{subfigure}
        \caption{Selected results of our GA-based solver for large puzzles. The first row shows: (a) 5,015-piece puzzle and the best chromosome achieved by the GA in the (b) first, (c) second, and (d) last generation. Similarly, the second and third rows show the best chromosomes in the same generations for 10,375- and 22,834-piece puzzles, respectively. The accuracy of all puzzle solutions shown is 100\%.}
        \label{fig:selectedSolutions}
\end{figure*}

\begin{table}[t]
\centering
\begin{tabular}{ |c||c|c|c||c| }
    \hline
  \# & Avg. & Avg. & Avg. & Avg. \\
  of Pieces  & Best & Worst & Avg. & Std. Dev. \\ \hline \hline
  432 & 96.16\% & 95.21\% & 95.70\% & 0.34\% \\ \hline
  540 & 95.96\% & 94.65\% & 95.38\% & 0.40\% \\ \hline
  805 & 96.26\% & 95.35\% & 95.85\% & 0.31\% \\ \hline
  2,360 & 88.86\% & 87.52\% & 88.00\% & 0.38\% \\ \hline
  3,300 & 92.76\% & 91.91\% & 92.37\% & 0.27\% \\ \hline
\end{tabular}
\caption{Accuracy results using our GA; set averages are given for each image set of best, worst, and average scores (as well as standard deviation) over 10 runs for each image.}
\label{tab:bestWorstAvgComp}
\end{table}

\begin{table}[t]
\centering
\begin{tabular}{ |c||c|c||c| }
    \hline
  \# of Pieces  & Pomeranz {\etal} & GA & Diff \\ \hline \hline
  432 & 94.25\% & 96.16\% & {\bf 1.91\%} \\ \hline
  540 & 90.90\% & 95.96\%  & {\bf 5.06\%} \\ \hline
  805 & 89.70\% & 96.26\% & {\bf 6.56\%} \\ \hline
  2,360 & 84.67\% & 88.86\% & {\bf 4.19\%} \\ \hline
  3,300 & 85.00\% & 92.76\% & {\bf 7.76\%} \\ \hline
\end{tabular}
\caption{Comparison of our accuracy results to those of Pomeranz {\etal} (derived from their supplementary material); averages are given for each image set of best scores over 10 runs for each image. }
\label{tab:pomAvgCompare}
\end{table}

\begin{table}[t]
\centering
\begin{tabular}{ |c||c|c||c| }
    \hline
  \# of Pieces  & Pomeranz {\etal} & GA & Diff \\ \hline \hline
  432 & 76.00\% & 81.06\% & {\bf 5.06\%} \\ \hline
  540 & 58.33\% & 79.32\% & {\bf 20.99\%} \\ \hline
  805 & 67.33\% & 86.30\% & {\bf 18.97\%} \\ \hline
  2,360 & 84.67\% & 88.86\% & {\bf 4.19\%} \\ \hline
  3,300 & 85.00\% & 92.76\% & {\bf 7.76\%} \\ \hline
\end{tabular}
\caption{Comparison of our accuracy results to those of Pomeranz {\etal}, relating only to the 3 least accurately solved images in every set.}
\label{tab:pomAvgWorseCompare}
\end{table}

Next, we have augmented the current benchmark by compiling three additional 20-image sets for this work and future studies. These image sets contain 5,015-, 10,375-, and 22,834-piece puzzles. (Note that the latter two puzzle sizes have never been solved before.) The results for the additional three sets are shown in Table~\ref{tab:largeAvg}. As before, we ran the GA 10 times on each image and recorded the best, worst, and average accuracy (as well as the standard deviation). Having observed little difference between the best and worst runs for puzzles with less than 22,834 pieces, and since this difference seemed to decrease for larger puzzles, we ran the GA only twice on each 22,834-piece puzzle. Even when challenged with an image size that has never been attempted, the GA functions well, achieving (almost) perfect results.

An interesting phenomenon observed is depicted in Figure~\ref{fig:shited}. For some puzzles, the GA managed to reach a "better-then-perfect" score, {\ie} a placement such that the dissimilarity is smaller than that of the original (correct) image. These placements were reproduced even when using more sophisticated metrics such as the one offered in~\cite{conf/cvpr/PomeranzSB11}. Moreover, in some cases, the correct solution was reached and then changed to the "better" one. As far as we know, observations of this kind were not documented before. Although undesirable, this manifestation is a proof of the GA's ability of reaching unprecedented accuracy levels. Obviously, revisiting the fitness function ({\ie} compatibility metric) would be required for these cases.

Finally, Table~\ref{tab:timesTable} depicts the average run time of the GA per image set; all experiments were conducted on a modern PC. When running on the smaller sets of 432, 540, and 805 pieces, the GA terminated after 48.73, 64.06, and 116.18 seconds, on average, respectively. These results are comparable to the times measured by Pomeranz {\etal}~\cite{conf/cvpr/PomeranzSB11} of 1.2 minutes, 1.9 minutes, and 5.1 minutes, on the same sets. Experimenting with the largest benchmark of 22,834-piece puzzles, the GA terminated, on average, after only 13.19 hours. In comparison, Gallagher~\cite{conf/cvpr/Gallagher12} reported a run time of 23.5 hours for solving a 9,600-pieces puzzle, although he allows also for piece rotation. In summary, when facing both smaller and larger images, the GA's performance seems to surpass previously reported performances achieved by other greedy algorithms.

\begin{table}[t]
\centering
\begin{tabular}{ |c||c|c|c||c| }
    \hline
  \# & Avg. & Avg. & Avg. & Avg. \\
  of Pieces  & Best & Worst & Avg. & Std. Dev. \\ \hline \hline
  5,015 & 95.25\% & 94.87\% & 95.06\% & 0.11\% \\ \hline
  10,375 & 98.47\% & 98.20\% & 98.36\% & 0.08\% \\ \hline
  22,834 & 96.28\% & 96.17\% & 96.22\% & 0.05\% \\ \hline

\end{tabular}
\caption{Accuracy results using our GA on larger images; set averages are given for each image set of best, worst, and average scores (as well as standard deviation) over 10 runs for each image (and 2 runs for each 22,834-piece puzzle).}
\label{tab:largeAvg}
\end{table}

\begin{table}[t]
\centering
\begin{tabular}{ |c||c|c|c||c| }
    \hline
  \# & Avg. & Avg. & Avg. & Avg. \\
  of Pieces  & Best & Worst & Avg. & Std. Dev. \\ \hline \hline
  432 & 86.19\% & 80.56\% & 82.94\% & 2.62\% \\ \hline
  540 & 92.75\% & 90.57\% & 91.65\% & 0.65\% \\ \hline
  805 & 94.67\% & 92.79\% & 93.63\% & 0.62\% \\ \hline
  2,360 & 85.73\% & 82.73\% & 84.62\% & 0.86\% \\ \hline
  3,300 & 89.92\% & 65.42\% & 86.62\% & 7.19\% \\ \hline
  5,015 & 94.78\% & 90.76\% & 92.04\% & 1.74\% \\ \hline
  10,375 & 97.69\% & 96.08\% & 97.12\% & 0.45\% \\ \hline
  22,834 & 92.02\% & 91.46\% & 91.74\% & 0.28\% \\ \hline

\end{tabular}
\caption{Results of running the GA 10 times on every image in every set under direct comparison; the best, worst, and average scores were recorded for every image.}
\label{tab:directAvg}
\end{table}

\begin{table}[t]
\centering
\begin{tabular}{ |c||c| }
    \hline
  \# of Pieces  & Run Time \\ \hline \hline
  432 & 48.73 [sec] \\ \hline
  540 &  64.06 [sec]\\ \hline
  805 & 116.18 [sec] \\ \hline
  2,360 & 17.60 [min] \\ \hline
  3,300 & 30.24 [min] \\ \hline
  5,015 & 61.06 [min] \\ \hline
  10,375 & 3.21 [hr] \\ \hline
  22,834 & 13.19 [hr] \\ \hline
\end{tabular}
\caption{Average run times of the GA on an image in every set.}
\label{tab:timesTable}
\end{table}

\section{Discussion and future work}

In this paper we presented an automatic jigsaw puzzle solver, far more accurate than any existing solver and capable of reconstructing puzzles of up to 22,834 pieces (more than twice the number of pieces ever achieved). We also created new sets of large images to be used for benchmark testing for this and other solvers, and supplied both the image sets and our results for the benefit of the community~\cite{conf/cvpr/site/Our}.

We have achieved the first effective genetic algorithm-based solver, which appears to challenge state-of-the-art performance of other jigsaw puzzle solvers. By introducing a novel crossover technique, we were able to arrive at an effective solver. Our approach could prove useful in future utilization of GAs for solving more difficult variations of the jigsaw problem (including unknown piece orientation, missing and excessive puzzle pieces, unknown puzzle dimensions, and three-dimensional puzzles), and could also assist in the design of GAs in other problem domains.



{\small
\bibliographystyle{ieee}
\bibliography{egbib}

\begin{thebibliography}{10}\itemsep=-1pt

\bibitem{journals/aai/Altman89}
T.~Altman.
\newblock Solving the jigsaw puzzle problem in linear time.
\newblock {\em Applied Artificial Intelligence an International Journal},
  3(4):453--462, 1989.

\bibitem{journals/tog/BrownTNBDVDRW08}
B.~Brown, C.~Toler-Franklin, D.~Nehab, M.~Burns, D.~Dobkin, A.~Vlachopoulos,
  C.~Doumas, S.~Rusinkiewicz, and T.~Weyrich.
\newblock A system for high-volume acquisition and matching of fresco
  fragments: Reassembling theran wall paintings.
\newblock {\em ACM Transactions on Graphics}, 27(3):84, 2008.

\bibitem{cao2010automated}
S.~Cao, H.~Liu, and S.~Yan.
\newblock Automated assembly of shredded pieces from multiple photos.
\newblock In {\em IEEE Int. Conf. on Multimedia and Expo}, pages 358--363,
  2010.

\bibitem{conf/cvpr/ChoAF10}
T.~Cho, S.~Avidan, and W.~Freeman.
\newblock A probabilistic image jigsaw puzzle solver.
\newblock In {\em IEEE Conference on Computer Vision and Pattern Recognition},
  pages 183--190, 2010.

\bibitem{bb43059}
T.~Cho, M.~Butman, S.~Avidan, and W.~Freeman.
\newblock The patch transform and its applications to image editing.
\newblock In {\em IEEE Conference on Computer Vision and Pattern Recognition},
  pages 1--8, 2008.

\bibitem{conf/icip/DeeverG12}
A.~Deever and A.~Gallagher.
\newblock Semi-automatic assembly of real cross-cut shredded documents.
\newblock In {\em ICIP}, pages 233--236, 2012.

\bibitem{springerlink:10.1007/s00373-007-0713-4}
E.~Demaine and M.~Demaine.
\newblock Jigsaw puzzles, edge matching, and polyomino packing: Connections and
  complexity.
\newblock {\em Graphs and Combinatorics}, 23:195--208, 2007.

\bibitem{bb47278}
H.~Freeman and L.~Garder.
\newblock Apictorial jigsaw puzzles: The computer solution of a problem in
  pattern recognition.
\newblock {\em IEEE Transactions on Electronic Computers}, EC-13(2):118--127,
  1964.

\bibitem{conf/cvpr/Gallagher12}
A.~Gallagher.
\newblock Jigsaw puzzles with pieces of unknown orientation.
\newblock In {\em IEEE Conference on Computer Vision and Pattern Recognition},
  pages 382--389, 2012.

\bibitem{GolMalBer04}
D.~Goldberg, C.~Malon, and M.~Bern.
\newblock A global approach to automatic solution of jigsaw puzzles.
\newblock {\em Computational Geometry: Theory and Applications},
  28(2-3):165--174, 2004.

\bibitem{holland1975adaptation}
J.~Holland.
\newblock Adaptation in natural and artificial systems, university of michigan
  press.
\newblock {\em Ann Arbor, MI}, 1(97):5, 1975.

\bibitem{justino2006reconstructing}
E.~Justino, L.~Oliveira, and C.~Freitas.
\newblock Reconstructing shredded documents through feature matching.
\newblock {\em Forensic science international}, 160(2):140--147, 2006.

\bibitem{journals/KollerL06}
D.~Koller and M.~Levoy.
\newblock Computer-aided reconstruction and new matches in the forma urbis
  romae.
\newblock {\em Bullettino Della Commissione Archeologica Comunale di Roma},
  pages 103--125, 2006.

\bibitem{journals/science/MarandeB07}
W.~Marande and G.~Burger.
\newblock Mitochondrial {DNA} as a genomic jigsaw puzzle.
\newblock {\em Science}, 318(5849):415--415, 2007.

\bibitem{marques2009reconstructing}
M.~Marques and C.~Freitas.
\newblock Reconstructing strip-shredded documents using color as feature
  matching.
\newblock In {\em ACM symposium on Applied Computing}, pages 893--894, 2009.

\bibitem{conf/ifip/MortonL68}
A.~Q. Morton and M.~Levison.
\newblock The computer in literary studies.
\newblock In {\em IFIP Congress}, pages 1072--1081, 1968.

\bibitem{conf/cvpr/PomeranzSB11}
D.~Pomeranz, M.~Shemesh, and O.~Ben-Shahar.
\newblock A fully automated greedy square jigsaw puzzle solver.
\newblock In {\em IEEE Conference on Computer Vision and Pattern Recognition},
  pages 9--16, 2011.

\bibitem{conf/cvpr/site/PomeranzSB11}
D.~Pomeranz, M.~Shemesh, and O.~Ben-Shahar.
\newblock A fully automated greedy square jigsaw puzzle solver {MATLAB} code
  and images. https://sites.google.com/site/greedyjigsawsolver/home, 2011.

\bibitem{conf/cvpr/site/Our}
D.~Sholomon, O.~David, and N.~Netanyahu.
\newblock Datasets of larger images and {GA}-based solver's results on these
  and other sets. http://www.cs.biu.ac.il/$\sim$nathan/{J}igsaw.

\bibitem{bb58987}
F.~Toyama, Y.~Fujiki, K.~Shoji, and J.~Miyamichi.
\newblock Assembly of puzzles using a genetic algorithm.
\newblock In {\em IEEE Int. Conf. on Pattern Recognition}, volume~4, pages
  389--392, 2002.

\bibitem{oai:xtcat.oclc.org:OCLCNo/ocm45147791}
C.-S.~E. Wang.
\newblock {\em Determining molecular conformation from distance or density
  data}.
\newblock PhD thesis, Massachusetts Institute of Technology, Dept. of
  Electrical Engineering and Computer Science, 2000.

\bibitem{yang2011particle}
X.~Yang, N.~Adluru, and L.~J. Latecki.
\newblock Particle filter with state permutations for solving image jigsaw
  puzzles.
\newblock In {\em IEEE Conference on Computer Vision and Pattern Recognition},
  pages 2873--2880. IEEE, 2011.

\bibitem{Zhao:2007:PSA:1348258.1348289}
Y.~Zhao, M.~Su, Z.~Chou, and J.~Lee.
\newblock A puzzle solver and its application in speech descrambling.
\newblock In {\em WSEAS Int. Conf. Computer Engineering and Applications},
  pages 171--176, 2007.

\end{thebibliography}
}

\end{document}